\documentclass[journal,onecolumn]{IEEEtran}

\newtheorem{theorem}{Theorem}[section]
\newtheorem{definition}{Definition}[section]
\newtheorem{lemma}{Lemma}[section]
\newtheorem{corollary}{Corollary}[section]
\newtheorem{remark}{Remark}[section]
\usepackage{enumerate}
\usepackage{bm}
\usepackage{diagbox}
\usepackage{color}
\usepackage{graphicx}
\usepackage{amssymb,amsmath}
\usepackage{ulem}
\usepackage{cancel}
\usepackage{cite}
\usepackage{subfigure,color}

\newtheorem{proposition}{Proposition}[section]

\begin{document}

\title{On Dimension-free Tail Inequalities for Sums of Random Matrices and Applications}

\author{Chao~Zhang~\IEEEmembership{Member,~IEEE,} Min-Hsiu~Hsieh$^*$~\IEEEmembership{Senior Member,~IEEE,} Dacheng Tao,~\IEEEmembership{Fellow,~IEEE}
\thanks{C.~Zhang is with the School of Mathematical Sciences, Dalian University of Technology, Dalian, Liaoning, 116024, P.R. China. E-mail: chao.zhang@dlut.edu.cn.}
\thanks{M.~Hsieh is with Centre for Quantum Software and Information, University of Technology Sydney, NSW 2007, Australia. E-mail: Min-Hsiu.Hsieh@uts.edu.au.}
\thanks{D.~Tao is with the School of Information Technologies, The University of Sydney, Darlington, NSW 2008, Australia. E-mail: dacheng.tao@sydney.edu.au.}
\thanks{This work is partially supported by the National Natural Science Foundation of China: 11401076 and 61473328.}
\thanks{$^*$Corresponding author}

}

\maketitle

\begin{abstract}

In this paper, we present a new framework to obtain tail inequalities for sums of random matrices. Compared with the existing works, the tail inequalities obtained under this framework have the following characteristics: 1) high feasibility --- they can be used to study the tail behavior of many kinds of matrix functions, {\it e.g.,} arbitrary kinds of matrix norms, the absolute value of the sum of the $j$ largest eigenvalues of Hermitian matrices and the sum of the $j$ largest singular values of complex matrices; and 2) independence of matrix dimension --- they do not have the matrix-dimension term as a product factor, and thus are suitable to the scenario of high-dimensional or infinite-dimensional random matrices. The price we pay to obtain these advantages is that the convergence rate of the resulting inequalities will become slow when the number of summand random matrices is large. However, this deficiency can be overcome by splitting these summand random matrices into fewer groups or further decreasing their magnitude. We also develop the tail inequalities for matrix random series, which are the sums of fixed matrices weighted by independent random variables, and matrix martingale difference sequence. We also demonstrate usefulness of our tail bounds in several fields. In compressed sensing, we employ the resulted tail inequalities to achieve a proof of the restricted isometry property (RIP) when the measurement matrix is the sum of random matrices without any assumption on the distributions of matrix entries. In probability theory, we derive a new upper bound to the supreme of stochastic processes. In machine learning, we prove new expectation bounds of sums of random matrices matrix and obtain matrix approximation schemes via random sampling. In quantum information, we show a new analysis relating to the fractional cover number of quantum hypergraphs. In theoretical computer science, we obtain randomness-efficient samplers using matrix expander graphs that can be efficiently implemented in time without dependence on matrix dimensions.

\end{abstract}

\begin{IEEEkeywords}
Random matrix, tail inequality, dimension-free, eigenvalue, singular value, restricted isometry property, compressed sensing, stochastic process, matrix approximation

\end{IEEEkeywords}

\section{Introduction}


\IEEEPARstart{O}{ne} major research topic on random matrices is to study the tail inequalities for sums of random matrices, which bound the probability of the extreme eigenvalues (or singular values) of sums of random matrices that are larger than a given constant ({\it cf.}\cite{ahlswede2002strong,hsu2012tail,minsker2017some,tropp2012user,vershynin2012,zhang2017lsv}). Random matrices have been widely used in many research fields, {\it e.g.,} compressed sensing \cite{vershynin2012}, quantum computing \cite{ahlswede2002strong} and optimization \cite{nemirovski2007sums,so2011moment}, to model practical system behaviours with uncertainty disturbance. Crucial system characteristics can be observed efficiently if the concentration phenomenon of the tail behaviour of the random fluctuation exists.  The following are some application examples of this study: 

\begin{itemize}
\item In compressed sensing, Baraniuk {\it et al.} \cite{baraniuk2008simple} introduced an alternative definition of the restricted isometric properties (RIP), and then achieve a simple proof for the RIP under the concentration assumption of the measurement matrix;

\item In optimization, Nemirovski \cite{nemirovski2007sums} and So \cite{so2011moment} have pointed out that the behavior of matrix random series, which is the sum of fixed matrices weighted by independent random variables, is strongly related to the efficiently computable solutions to many optimization problems, {\it e.g.,} the chance constrained optimization problem and the quadratic optimization problem with orthogonality constraints; 

\item In probability theory, Hsu {\it et al.} \cite{hsu2011dimension} used the tail inequalities of random matrices to bound the supremum of stochastic processes;

\item In machine leaning, the tail inequalities have been applied to study matrix approximation via random sampling \cite{tropp2015introduction}.

\item {In theoretical computer science, the matrix Chernoff bounds have been used to prove that the matrix-valued samples taken from the stationary random walk on an expander graph can most be treated as the independent samples \cite{wigderson2005randomness,wigderson2008derandomizing,kyng2018matrix,garg2018matrix}. }

\item {In quantum information, quantum systems are naturally in matrix forms, and hence tail bounds are very useful to study the quantum system behaviours with random noise \cite{ahlswede2002strong}. }

\end{itemize}


Most existing tail inequalities are equipped with the matrix-dimension term as a product factor, and thus are unsuitable to the scenario of high-dimensional or infinite-dimensional matrices. To overcome this shortcoming, it is important yet challenging to develop the dimension-free tail inequalities for sums of random matrices. Instead of the ambient matrix dimension, some pioneering works introduced the intrinsic matrix dimension to reduce the matrix-dimension dependence in the tail inequalities for sums of random matrices ({\it cf.} \cite{minsker2017some,hsu2012tail,tropp2015introduction}). Moreover, Rudelson and Vershynin \cite{rudelson2007sampling} presented the dimension-free tail inequalities for the sum of rank-one random matrices, each of which can be expressed as the tensor product of a bounded random vector with itself. Recently, Hsu {\it et al.} \cite{hsu2011dimension} gave the tail results for sums of random matrices by replacing the explicit matrix dimensions with a trace quantity that could be small when the matrix dimension is large or infinite. Magen and Zouzias \cite{magen2011low} applied the non-commutative Khintchine moment inequality to achieve a dimension-free tail inequality for sums of low-rank bounded matrices while the convergence rate of this inequality will be slow because of the absence of exponential form.


Consider the sum of $K$ random Hermitian matrices ${\bf X}_1,\cdots,{\bf X}_K\in\mathbb{C}^{n\times n}$, ${\bf Y}:=\sum_{k=1}^K {\bf X}_k$. In the literature, obtaining the tail inequalities for the largest eigenvalue of ${\bf Y}$ has a common starting-point ({\it cf.} \cite{ahlswede2002strong,hsu2012tail,minsker2017some,tropp2012user}): 
\begin{equation}\label{eq:tropp.laplace}
\mathbb{P}\{\lambda_{\max}({\bf Y})\geq t\}\leq \inf_{\theta>0}\big\{{\rm e}^{-\theta t}\cdot \mathbb{E}\,{\rm tr}\,{\rm e}^{\theta {\bf Y}}\big\},\quad t>0,
\end{equation}
where $\lambda_{\max}(\cdot)$ denotes the largest eigenvalue.
This suggests that the key to bounding $\mathbb{P}\{\lambda_{\max}({\bf Y})\geq t\}$ is to obtain the relevant Laplace-transform bounds, {\it i.e.,} the upper bound of $\mathbb{E}\,{\rm tr}\,{\rm e}^{\theta {\bf Y}}$ ({\it cf.} \cite[Proposition 3.1]{tropp2012user}). By using the Golden-Thompson trace inequality, Ahlswede and Winter \cite{ahlswede2002strong} arrived at
\begin{eqnarray}\label{eq:aw}
\mathbb{E}\,{\rm tr}\,{\rm e}^{\theta {\bf Y}}={\rm tr}\,\mathbb{E}\,{\rm e}^{\theta {\bf Y}}
&\leq& ({\rm tr}{\bf I})\cdot \Big[\prod_k \lambda_{\max}(\mathbb{E}{\rm e}^{\theta {\bf X}_k})  \Big] \\
&=&{\rm dim}({\bf Y})\cdot \exp\Big(\sum_k\lambda_{\max}
\big(\log \mathbb{E} {\rm e}^{\theta {\bf X}_k}   \big) \Big), \label{eq:aw01}
\end{eqnarray}
where ${\rm dim}({\bf Y})$ stands for the matrix dimension, or the ambient dimension (AD), of ${\bf Y}$. Note that there are two shortcomings in the bound: 
\begin{enumerate}[(i)]
\item The form of $\sum$-over-$\lambda_{\max}$ on the right-hand side of (\ref{eq:aw01}) is potentially much larger than the optimal  form of $\lambda_{\max}$-over-$\sum$.

\item The inequality has the matrix dimension ${\rm dim}({\bf Y})$ as a product factor; hence, it will become very loose when the dimension of ${\bf Y}$ is high.
\end{enumerate}

The first shortcoming has been successfully solved in Tropp's work \cite{tropp2012user}, where Lieb's concavity theorem was applied to obtain the following Laplace-transform bound \cite[Lemma 3.4]{tropp2012user}:
\begin{equation}\label{eq:tropp}
\mathbb{E}\,{\rm tr}\,{\rm e}^{\theta {\bf Y}}\leq {\rm dim}({\bf Y})\cdot
\exp\left(\lambda_{\max}\left(\sum_k\log \mathbb{E}{\rm e}^{\theta {\bf X}_k}    \right)\right).
\end{equation}
Denote $v = \lambda_{\max}(\sum_k\mathbb{E} {\bf X}_k^2)$, and $\lambda_{\max}({\bf X}_k) \leq L$ for all $k$. One can obtain the tail inequality \cite[Theorem 6.1]{tropp2012user}):\footnote{The inequality of 
\eqref{eq:amb.tail01} is resulted from the fact that $\Gamma(t)\leq \frac{t^2}{2(1+t/3)}$ when $t>0$.}
\begin{eqnarray}\label{eq:amb.tail}
\mathbb{P}\big\{\lambda_{\max}({\bf Y})\geq t \big\} &\leq&  {\rm dim}({\bf Y})\cdot \exp \left(  -\frac{v}{L^2}\cdot \Gamma\left( \frac{Lt}{v} \right) \right) \\
 &\leq&  {\rm dim}({\bf Y})\cdot \exp \left(  \frac{-t^2/2}{v+Lt/3} \right), \label{eq:amb.tail01}
\end{eqnarray}
where $t>0$ and
\begin{equation}\label{eq:gamma}
\Gamma(t):= (t+1)\log(t+1)-t.
\end{equation}

The introduction of the intrinsic dimension (ID) $${\rm intdim}({\bf V}):=\frac{{\rm tr}({\bf V})}{\lambda_{1}({\bf V})},$$ where $\sum_k\mathbb{E} \{{\bf X}_k {\bf X}_k^*\} \preceq {\bf V} $ with ${\bf A} \preceq {\bf B}$ meaning that the matrix ${\bf B}-{\bf A}$ is a positive semidefinite matrix   \cite{hsu2012tail,minsker2017some,tropp2015introduction}, provides an attempt to overcome the second shortcoming. By setting $\Psi(t) := {\rm e}^{\theta t} - \theta t -1$ in the generalized Laplace transform bound ({\it cf.} \cite[Proposition 7.4.1]{tropp2015introduction}):
\begin{equation*}
\mathbb{P}\big\{\lambda_{\max}({\bf Y})\geq t \big\} \leq \frac{1}{\Psi(t)} \cdot \mathbb{E} \, {\rm tr}\, \Psi({\bf Y}),
\end{equation*}
Tropp obtained the tail inequality with the intrinsic dimension as a product factor  \cite[Theorem 7.3.1]{tropp2015introduction}): 
\begin{equation}\label{eq:int.tail}
\mathbb{P}\big\{\lambda_{\max}({\bf Y})\geq t \big\} \leq 4\cdot {\rm intdim}({\bf V})\cdot \exp \left(  \frac{-t^2/2}{v+Lt/3} \right),
\end{equation}
for $t\geq \sqrt{v} + L/3$.
Although ${\rm intdim}({\bf V})$ could be smaller than ${\rm dim}({\bf Y})$, there is still one caveat for this method, namely,
 the ID inequality \eqref{eq:int.tail} cannot be used to study $\mathbb{P}\big\{\lambda_{\max}({\bf Y})\geq t \big\}$ when $t$ is smaller than $\sqrt{v} + L/3$. Therefore, it is desirable to obtain the tail inequalities for sums of random matrices without the aforementioned shortcomings. 



\subsection{Overview of Main Results}

In this paper, we propose a new framework to obtain the dimension-free tail inequalities for sums of random matrices. Compared with the existing works, the tail inequalities obtained under this framework has the following advantages. 
\begin{enumerate}[(i)]

\item They can be used to study the tail behavior of several kinds of matrix functions including arbitrary kinds of matrix norms, the sum of the $j$ largest singular values for complex matrices and the absolute value of the sum of the $j$ largest eigenvalues for Hermitian matrices.
\item The tail bounds do not have a dimensional factor, and can work for any $t>0$. Because of the independence of the matrix dimension, they are suitable to the scenario of high-dimensional or infinite-dimensional matrices. 

\end{enumerate}

Under this framework, we further obtain the tail inequalities for matrix random series, which are the sums of fixed matrices weighted by independent random variables. Compared with the existing works \cite{tropp2012user,zhang2018matrix}, our results are independent of the matrix dimension but also suitable to arbitrary kinds of probability distributions with bounded first-order moment.

As an application in compressed sensing, we apply the resulted tail inequalities to achieve a proof of the RIP of the measurement matrix that can be expressed as sums of random matrices without any assumption imposed on the entries of matrices. In addition, we also discuss the applications of our results in optimization, stochastic process, matrix approximation, {and quantum information.}

The rest of this paper is organized as follows. In Section \ref{sec:notation}, we introduce some necessary preliminaries and notations. In Section \ref{sec:main}, we present the main results of this paper. In Section \ref{sec:rip}, we discuss the application of the resulted tail inequalities in compressed sensing. {The applications in stochastic process, matrix approximation via random sampling and quantum information are given in Section \ref{sec:process}, Section \ref{sec:approximation} and Section \ref{sec:sampling}, respectively.} The last section concludes the paper and the proofs of our main results are given in the appendix.

iiii
\section{Preliminaries and Notations}\label{sec:notation}

In this section, we introduce some necessary preliminaries and notations, and then give a Laplace-transform bound as the starting-point to obtain the main results.

\subsection{Matrix Functions}

Let $\mu: \mathbb{M}\rightarrow \mathbb{R}$ be a matrix function defined on the matrix set $\mathbb{M}$. Assume that the function $\mu$ and the matrix set $\mathbb{M}$ satisfy the following conditions: 
\begin{description}
\item[(C1)] For any ${\bf A}\in\mathbb{M}$, it holds that $\mu({\bf A}) \geq 0$.
\item[(C2)] For any ${\bf A}\in\mathbb{M}$ and any $\theta\geq 0$, it holds that $\theta\cdot {\bf A} \in\mathbb{M}$ and $\mu(\theta\cdot {\bf A}) = \theta \cdot \mu( {\bf A})$.
\item[(C3)] For any ${\bf A},{\bf B}\in\mathbb{M}$, it holds that $\mu({\bf A}+{\bf B})\leq \mu({\bf A})+\mu({\bf B})$.
\end{description}
The following are examples of the function $\mu(\cdot)$ and the matrix set $\mathbb{M}$ satisfying conditions (C1)-(C3):
\begin{enumerate}[(i)]
\item According to \cite[Corollary 3.4.3]{horn1991topics}, the function $\mu(\cdot)$ can be the sum of the $j$, $1\leq j < \min\{m,n\}$, largest singular values $\sum_{i=1}^j\sigma_{i}(\cdot)$ in the case that $\mathbb{M} = \mathbb{C}^{m\times n}$, where the notation $\mathbb{C}^{m\times n}$ stands for the set of all $m\times n$ complex matrices and all singular values $\sigma_1,\cdots,\sigma_{\min\{m,n\}}$ are in descending order, {\it i.e.}, $\sigma_1\geq \cdots\geq\sigma_{\min\{m,n\}}$.
\item According to \cite[Theorem G.1]{marshall2010inequalities}, the function $\mu(\cdot)$ can be the absolute value of the sum of the $j$, $1\leq j < n$, largest eigenvalues $\big|\sum_{i=1}^j\lambda_{i}(\cdot)\big|$ in the case that $\mathbb{M} = \mathbb{H}^{n\times n}$, where the notation $\mathbb{H}^{n\times n}$ denotes the set of all $n$-dimensional Hermitian matrices and all eigenvalues $\lambda_1,\cdots,\lambda_n$ are in descending order, {\it i.e.}, $\lambda_1\geq \cdots\geq\lambda_n$.

%

\item {It follows from the non-negativity, the homogeneousness and the triangle inequality that the function $\mu(\cdot)$ can be an arbitrary matrix norm with $\mathbb{M} = \mathbb{C}^{m\times n}$.}
\end{enumerate}

{Note that all results in this paper are built under the assumption that the function $\mu:\mathbb{M}\rightarrow \mathbb{R}$ and the matrix set $\mathbb{M}$ satisfy conditions (C1)-(C3) if no specific statements are given.}


\subsection{Infinite-dimensional Diagonal Matrices}

For any $\theta>0$, define an infinite-dimensional diagonal matrix w.r.t. the function $\mu$:
\begin{align}\label{eq:def.diag}
\widehat{{\bf D}}_\mu[\theta; {\bf B}]:={\bf D}_0+ {{\bf D}}_\mu[\theta; {\bf B}]
\end{align}
with
\begin{align}\label{eq:def.diag0}
{\bf D}_0:=&\bm{\Lambda}\left[0,0,\log\frac{1}{2!}, \log\frac{1}{3!},\cdots\right]
\end{align}
and
\begin{align}\label{eq:def.diag1}
{{\bf D}}_\mu&[\theta; {\bf B}]:= \bm{\Lambda}\Big[0, \log(\theta \cdot \mu({\bf B})+1),2\log(\theta \cdot \mu({\bf B})+1), 3\log(\theta \cdot \mu({\bf B})+1),\cdots\Big],
\end{align}
where $\bm{\Lambda}[ \cdots]$ stands for the diagonal matrix. It is direct that ${\rm tr} \,{\rm e}^{{\bf D}_0} = {\rm e}$. Subsequently, we consider the following properties of the operation ${\bf D}_\mu[\cdot\,;\,\cdot]$:
%
\begin{proposition}\label{prop:operation}
Given $K$ fixed matrices ${\bf B}_1,\cdots,{\bf B}_K\in \mathbb{M}$, let $\Omega = \{\Omega_1,\cdots,\Omega_I\}$ be a partition of the index set $\{1,\cdots, K\}$ with $\bigcup_{i=1}^I \Omega_i=\{1,\cdots, K\}$ and $|\Omega_i|$ stands for the cardinality of the set $\Omega_i$. Then, there holds that
\begin{align}
{\bf D}_\mu\left[\theta;\sum_{k=1}^K{\bf B}_k\right] \preceq \sum_{k=1}^K{\bf D}_\mu[\theta;{\bf B}_k];&\qquad\mbox{(sub-additivity)}\label{eq:operation1}\\
\sum_{k=1}^K {\bf D}_\mu\left[\theta;{\bf B}_k\right] \preceq K\cdot {\bf D}_\mu\left[\theta;\sum_{k=1}^K\frac{{\bf A}_k}{K}\right];&\qquad\mbox{(super-additivity)}\label{eq:operation2}\\
\sum_{k=1}^K{\bf D}_\mu[\theta;{\bf B}_k]  \preceq   \sum_{i=1}^I \left( |\Omega_i |\cdot  {\bf D}_\mu\left[\theta;\sum_{k\in\Omega_i} \frac{{\bf A}'_k}{|\Omega_i|}\right] \right),&\qquad \mbox{(partial super-additivity)}\label{eq:operation3}
\end{align} 
where ${\bf A}_1,\cdots,{\bf A}_K \in\mathbb{M}$ and ${\bf A}'_1,\cdots,{\bf A}'_K \in\mathbb{M}$ satisfy that $\sum\limits_k\mu({\bf B}_k)\leq\mu\big(\sum\limits_k{\bf A}_k\big)$ and $\sum\limits_{k\in\Omega_i}\mu({\bf B}_k)\leq\mu\big(\sum\limits_{k\in\Omega_i}{\bf A}'_k\big)$, respectively.

 \end{proposition}
In this proposition, the sub-additivity is a direct result from (C3), and the super-additivity (or partial super-additivity) can be achieved by using the inequality of Arithmetic and geometric means: $\frac{s_1+s_2+\cdots+s_K}{K} \geq \sqrt[K]{s_1  s_2  \cdots  s_K}$, $\forall \,s_1,\cdots,s_K \geq 0$. We remark that a simple choice of the matrices ${\bf A}_k$ and ${\bf A}'_k$ is as follows:
\begin{align}\label{eq:notationU}
{\bf A}_k\leftarrow& {{\bf U}}: = \mathop{\arg\max}\limits_{1\leq k\leq K}\big\{ \mu({\bf B}_k)\big\},\quad k\in\{1,2,\cdots, K\}\nonumber\\
{\bf A}'_k\leftarrow& {{\bf U}_i}: = \mathop{\arg\max}\limits_{k\in\Omega_i}\big\{ \mu({\bf B}_k)\big\},\quad k\in \Omega_i.
\end{align}


\subsection{Laplace-Transform Bounds}

Here, we use the aforementioned infinite-dimensional diagonal matrices ${\bf D}_\mu[\cdot;\cdot]$ to obtain the Laplace-transform bounds for the matrix function $\mu$. This provides a starting-point to achieve the tail inequalities for sums of random matrices. 

\begin{proposition}\label{prop:diagonal}
For any ${\bf B}\in\mathbb{M}$ and $\theta>0$,
\begin{equation}\label{eq:diagonal.1}
{\rm e}^{\mu(\theta \cdot {\bf B})}={\rm e}^{-1}\cdot {\rm tr}\,{\rm e}^{\widehat{{\bf D}}_\mu[\theta; {\bf B}]},
\end{equation}
and for any $K\in\mathbb{N}$,
\begin{equation}\label{eq:diagonal.2}
{\rm e}^{(\mu(\theta \cdot {\bf B})+1)^K}={\rm tr}\,{\rm e}^{{\bf D}_0+K\cdot {\bf D}_\mu[\theta; {\bf B}]},
\end{equation}
where $\widehat{{\bf D}}_\mu[\theta; {\bf B}]$ is defined in \eqref{eq:def.diag}.
\end{proposition}
These results are derived from the Taylor's expansion of ${\rm e}^{x}$, where the function $\mu(\theta {\bf X})$ is converted into the trace operation for the infinitely-dimensional diagonal matrix ${\bf D}_\mu[\theta;{\bf X}]$. Next, we consider the Laplace-transform bound for sums of random matrices, {\it i.e.,} the upper bound of $\mathbb{E} \,{\rm e}^{\theta \mu( \sum_k{\bf X}_k)}$.

\begin{proposition}\label{prop:diagonal.sum}
Let ${\bf X}_1,\cdots,{\bf X}_K\in \mathbb{M}$ be independent random matrices. Then, it holds that for any $\theta>0$,
\begin{align}\label{eq:diagonal.sum}
\mathbb{E}{\rm e}^{\mu\left(\sum_{k=1}^K\theta {\bf X}_k\right)}\leq {\rm e}^{-1}\cdot {\rm tr}\,\exp\left({\bf D}_0+\sum_{k=1}^K \log\mathbb{E}\,{\rm e}^{{\bf D}_\mu[\theta;{\bf X}_k]}\right).
\end{align}
\end{proposition}
%
This bound follows from the sub-additivity of ${\bf D}_\mu[\cdot\,;\,\cdot]$. Note that the term $\sum_{k=1}^K \log\mathbb{E}\,{\rm e}^{{\bf D}_\mu[\theta;{\bf X}_k]}$ in the right-hand side of \eqref{eq:diagonal.sum} could be improved to the desired form $ \log\mathbb{E}\,{\rm e}^{{\bf D}_\mu[\theta;\sum_k{\bf X}_k]}$. This can be done using the super-additivity of ${\bf D}_\mu[\cdot\,;\,\cdot]$. The detail is provided in the next section.


\section{Main Results}\label{sec:main}

In this section, we present the dimension-free (DF) tail inequalities for sums of random matrices. We also compare our results with the dimension-dependent bounds \eqref{eq:amb.tail} and \eqref{eq:int.tail}, and obtain a trade-off relationship between the matrix dimension and the number $K$ of the summand matrices.  We then provide numerical experiments 
and show that our tail inequalities yield a more accurate description to the tail behavior of sums of random matrices in some cases.

%


\subsection{Dimension-free Tail Inequalities}

Let $g(\theta,K)$ be an arbitrary function satisfying $g(\theta,K)\geq \max\{\theta,\theta^K\}$, $\forall\,\theta >0$. Given independent random matrices ${\bf X}_1,\cdots,{\bf X}_K\in \mathbb{M}$, let ${\bf B}_1,\cdots,{\bf B}_K\in\mathbb{M}$ be fixed matrices such that
\begin{equation}\label{eq:cond.b}
\mathbb{E}\,{\rm e}^{{\bf D}_\mu[\theta;{\bf X}_k]} \preceq {\rm e}^{{\bf D}_\mu[\theta;{\bf B}_k]},\quad 1\leq k\leq K.
\end{equation}
Denote $\phi:=\left[\mu\left({\bf U}\right)+1\right]^K-1$ with ${\bf U} = \mathop{\arg\max}\limits_{1\leq k\leq K} \{\mu({\bf B}_k)\}$. Then, we obtain the following master tail inequality for sums of random matrices.

\begin{figure}[htbp]
\centering
\subfigure[\hbox{The curves of $g_1(\theta,K)$ and $\max\{ \theta ,\theta^K\}$ ($K=2$).}]{
\includegraphics[height=6cm]{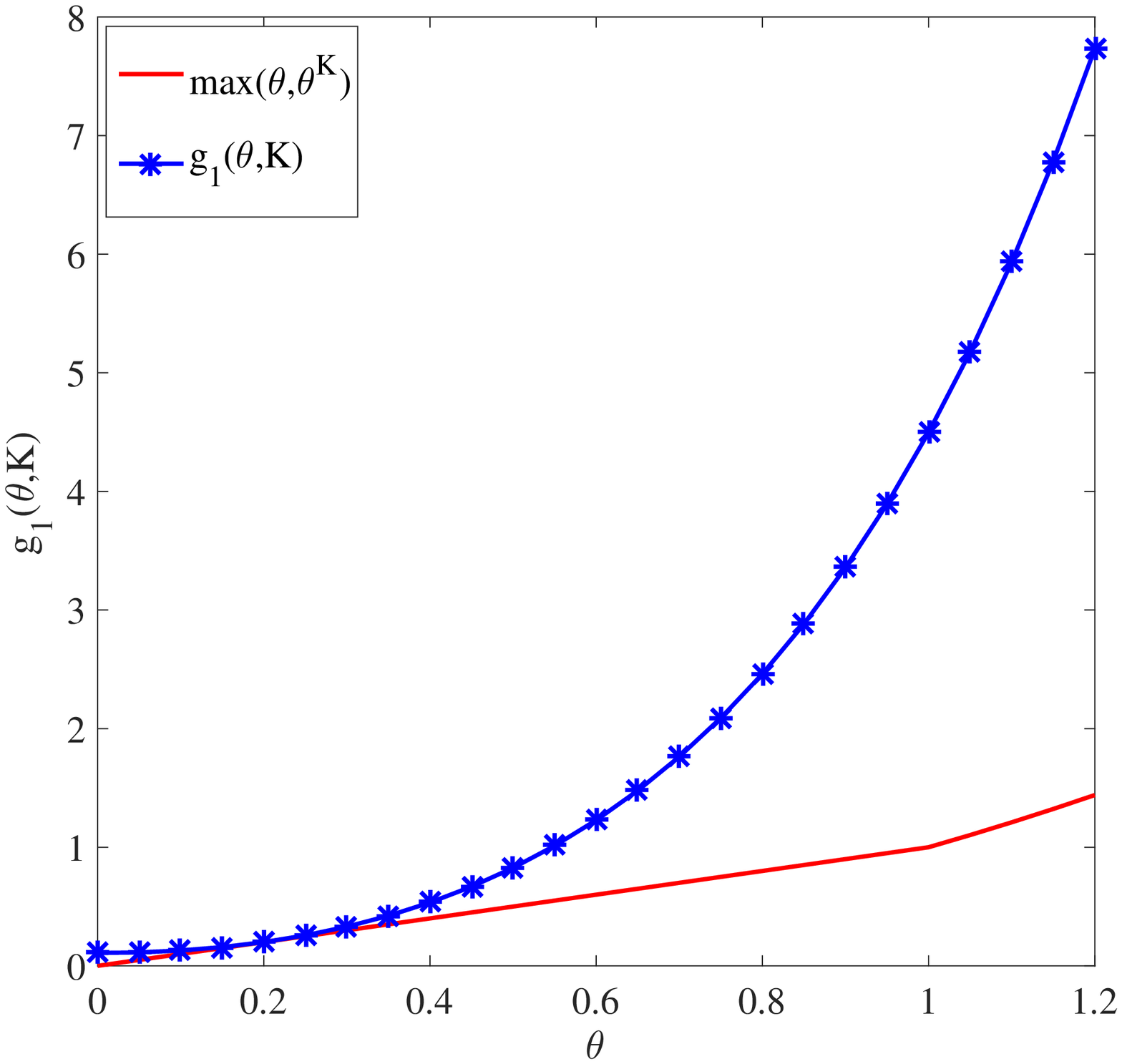} }
\subfigure[\hbox{The curve of $\alpha_1(K)$.}]{
\includegraphics[height=6cm]{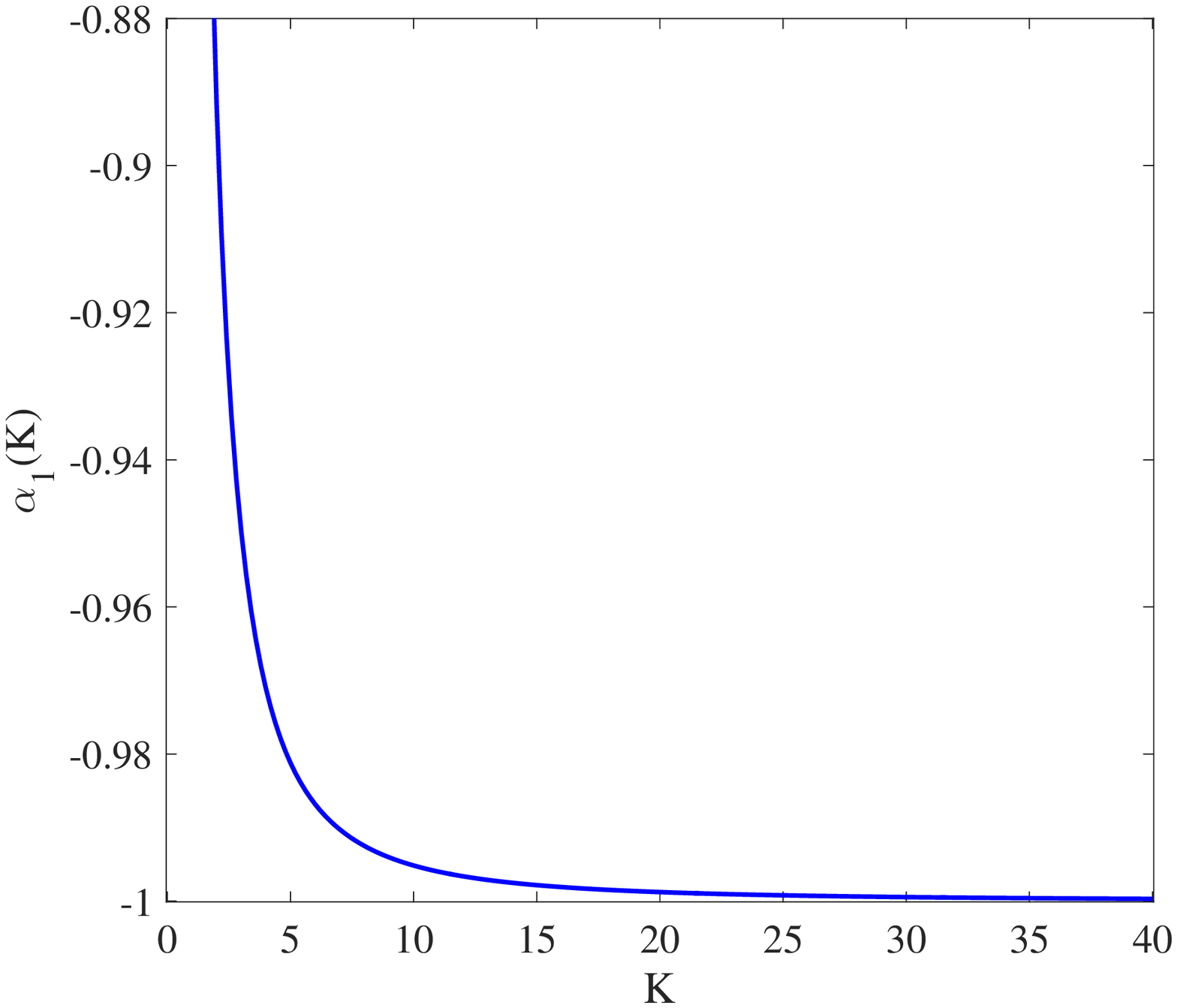}}
\caption{The function curves of $g_1(\theta,K)$ when $K=2$ and $\alpha_1(K)$.}\label{fig:g1}
\end{figure}

\begin{proposition}\label{prop:master1}
Let  Then, it follows that, for any $t>0$,
\begin{equation}\label{eq:master1}
\mathbb{P}\left\{\mu\left( \sum_k{\bf X}_k\right)\geq t  \right\}
\leq \inf_{\theta>0}\left\{{\rm e}^{-\theta t+g(\theta,K)\cdot \phi}  \right\}.
\end{equation}
\end{proposition}
%
As shown in the proof of this proposition, we first introduce fixed matrices ${\bf B}_1,\cdots,{\bf B}_K$ to control the behavior of the  random matrices $\{{\bf X}_k\}$, and then relax the bound to the one with the term $\sum_{k=1}^K {\bf D}_\mu[\theta;{\bf B}_k]$. Based on the super-additivity \eqref{eq:operation2} of ${\bf D}_\mu[\cdot;\cdot]$, we finally obtain the bound incorporating the term $\phi$. Note that 
one feasible choice of $g(\theta,K)$ is 
\begin{equation}\label{eq:g}
g(\theta,K) \leftarrow g_1(\theta,K):= {\rm e}^{K \theta} -K \theta + \alpha_1(K),
\end{equation}
where
\begin{equation}\label{eq:alpha}
\alpha_1(K):= \frac{K+1}{K}\left( \log \left(\frac{K+1}{K}\right) -1 \right).
\end{equation}
The function $g_1(\theta,K)$ is tangent to $\theta$ at the point $\big(\frac{1}{K}\log \frac{K+1}{K} , \frac{1}{K} \log \frac{K+1}{K} \big)$. Figure \ref{fig:g1} plots the curve of $g_1(\theta,K)$ when $K=2$ and $\alpha_1(K)$. 

By substituting $g_1(\theta,K)$ into the master inequality \eqref{eq:master1} and then taking minimization w.r.t. $\theta$, we arrive at the following result.
\begin{theorem}\label{thm:tail1}
For any $t>0$, then it holds that
\begin{align}\label{eq:tail1}
\mathbb{P}\left\{\mu\left(\sum_{k=1}^K{\bf X}_k \right)\geq t \right\}\leq& {\rm e}^{(1+\alpha_1(K))\cdot \phi}\cdot\exp\left(- \phi\cdot \Gamma\left(\frac{t}{K\phi}\right)\right) \nonumber\\
\leq& {\rm e}^{(1+\alpha_1(K))\cdot \phi}\cdot\exp\left(-   \frac{t^2/2}{K^2 \phi + Kt/3}\right)\nonumber\\
 \leq&
\left\{
\begin{array}{ll}
  {\rm e}^{\phi(1+\alpha_1(K)) }\cdot {\rm e}^{\frac{-t^2}{ 4K^2 \phi }} , &  \mbox{if $t< 3K\phi $;}    \\
   {\rm e}^{\phi(1+\alpha_1(K)) }\cdot {\rm e}^{\frac{-3t}{ 4K }}, &    \mbox{if $t\geq 3K\phi $,}   
\end{array}
\right.
\end{align}
where $\Gamma(t)$ and $\alpha_1(K)$ are defined in \eqref{eq:gamma} and \eqref{eq:alpha}, respectively.
\end{theorem}



Because of the appearance of the function $\Gamma(t)$, this result has the similar form of the dimension-dependent tail inequalities \eqref{eq:amb.tail} and \eqref{eq:int.tail}. However,  it also has the following different characteristics.
\begin{enumerate}
\item It does not have the matrix dimensional term as a product factor, such as ${\rm dim} ({\bf Y})$ (resp. ${\rm intdim} ({\bf Y})$) in \eqref{eq:amb.tail} (resp. \eqref{eq:int.tail}).

\item The function $\mu(\cdot)$ can be chosen as the sum of the $j$ largest singular values for complex matrices or the absolute value of the sum of the $j$ largest eigenvalues for Hermitian matrices. However, the inequalities \eqref{eq:amb.tail} and \eqref{eq:int.tail} are designed for the largest eigenvalue of  Hermitian matrices only.

\item There is no restriction on the random matrices ${\bf X}_k$ except that $\mathbb{E}\,{\rm e}^{{\bf D}_\mu[\theta;{\bf X}_k]} \preceq {\rm e}^{{\bf D}_\mu[\theta;{\bf B}_k]}$. In contrast, the inequalities \eqref{eq:amb.tail} and \eqref{eq:int.tail} require that the largest eigenvalues of ${\bf X}_k$ {are bounded.}

\item Since the summand number $K$ and the term $\phi$, which is in the order of $O[(1+\mu({\bf U}))^K]$, appear in the denominator of the right-hand side of \eqref{eq:tail1}, a large $K$ can bring a low rate of convergence to {\it zero} as $t$ goes to the {\it infinity}. In contrast, the convergence rate of the dimension-dependent inequalities \eqref{eq:amb.tail} and \eqref{eq:int.tail} are insensitive to the value of $K$. 

\end{enumerate}
To sum up, the tail inequality \eqref{eq:tail1} does not have the matrix dimension as a product factor, and thus could be suitable to studying the tail behavior of many spectral problems for high-dimensional (or infinite-dimensional) random matrices. However, since its convergence rate is sensitive to the value of $K$, it will converge to {\it zero} at a slow rate as $t$ goes to the {\it infinity} in the case of large $K$. 
To address this issue, we employ the partial super-additivity \eqref{eq:operation3} to obtain a variant of the term $\phi$ that has a lower order than $O[(1+\mu({\bf U}))^K]$. 

Let $\Omega = \{\Omega_1,\cdots,\Omega_I\}$ be a partition of the index set $\{1,\cdots, K\}$ with $\bigcup_{i=1}^I \Omega_i=\{1,\cdots, K\}$ and $\tau: = \max\limits_{1\leq i\leq I}\{ |\Omega_i|\}$. Denote $\phi_\Omega:=\sum\limits_{i=1}^I\big( \big[\mu\big({\bf U}_i\big)+1\big]^{|\Omega_i|}-1\big)$ with ${\bf U}_i = \mathop{\arg\max}\limits_{k\in\Omega_i} \{\mu({\bf B}_k)  \}$. Then, we get the following master tail inequality:

\begin{proposition}\label{prop:master2}
For any $t>0$, it holds that
\begin{equation}\label{eq:master.bound2}
\mathbb{P}\left\{\mu\left( \sum_k{\bf X}_k\right)\geq t  \right\}
\leq \inf_{\theta>0} \exp\left( -\theta t+ g(\theta,\tau) \cdot \phi_\Omega  \right).
\end{equation}
\end{proposition}
Compared with the tail bound \eqref{eq:master1}, this bound (\ref{eq:master.bound2}) incorporates the term $\phi_\Omega$ instead of the ordinary term $\phi$, which probably becomes explosive when $K$ is large because of the power $K$. In contrast, if the partition $\Omega = \{\Omega_1,\cdots,\Omega_I\}$ is well designed, the variant $\phi_\Omega$ will have a relatively lower power {$\tau$} and then the value of $\phi_\Omega$ will be controlled. In the similar way to achieve Theorems \ref{thm:tail1}, substituting $g(\theta,\tau):={\rm e}^{\tau\theta}-\tau\theta -\alpha_1(\tau) \geq \max\{\theta, \theta^{\tau}\}$ into the above master tail inequality leads to the following tail result.

\begin{theorem}\label{thm:tail2}
For any $t>0$, it holds that
\begin{align}\label{eq:tail2}
\mathbb{P}\left\{\mu\left(\sum_{k=1}^K{\bf X}_k \right)\geq t \right\}\leq& {\rm e}^{\phi_\Omega(1+\alpha_1(\tau))}\cdot\exp\left(- \phi_\Omega\cdot \Gamma\left(\frac{t}{\tau\cdot \phi_\Omega}\right)\right)\nonumber\\
 \leq& {\rm e}^{\phi_\Omega(1+\alpha_1(\tau))}\cdot\exp\left(-   \frac{t^2/2}{\tau^2 \phi_\Omega + \tau\cdot t/3}\right)\nonumber\\
 \leq&
\left\{
\begin{array}{ll}
  {\rm e}^{\phi_\Omega(1+\alpha_1(\tau)) }\cdot {\rm e}^{\frac{-t^2}{ 4\tau^2 \phi_\Omega }} , &  \mbox{if $t< 3\tau\phi_\Omega $;}    \\
   {\rm e}^{\phi_\Omega(1+\alpha_1(\tau)) }\cdot {\rm e}^{\frac{-3t}{ 4\tau }}, &    \mbox{if $t\geq 3\tau\phi_\Omega $.}   
\end{array}
\right.
\end{align}
\end{theorem}

{The number $\tau$ (resp. the term $\phi_\Omega$) appearing in the denominator of the right-hand side of \eqref{eq:tail2} is smaller than the number $K$ (resp. the term $\phi$) appearing in \eqref{eq:tail1}.} Therefore, compared with the aforementioned inequality \eqref{eq:tail1}, this inequality is less sensitive to the value of $K$ and converges to {\it zero} much slower as $t$ goes to the {\it infinity}. 

\begin{remark}\label{rem:suggestion}
Selecting the partition $\Omega = \{\Omega_1,\cdots,\Omega_I\}$ of the index set $\{1,2,\cdots,K\}$ plays an essential part in the practical application of this tail result. To control the order of the term $\phi_\Omega$, the partition $\Omega$ could be chosen in the following way:
\begin{itemize}
\item if $K$ is even, let each element of $\Omega$ contain two indexes, {\it i.e.,} $I = K/2$ and $\tau=2$;

\item {if $K$ is odd, one element of $\Omega$ contains one index and each of the others is composed of two indexes, {\it i.e.,} $I = (K+1)/2$ and $\tau=2$. }
\end{itemize}
{In the following discussion, such a partition is denoted as $\widetilde{\Omega} =  \{\widetilde{\Omega}_1,\cdots,\widetilde{\Omega}_{\widetilde{I}} \}$ with $\widetilde{I} = \lceil \frac{K}{2}  \rceil$}.
\end{remark}

\begin{remark}\label{rem:number}
One key difference between the resulted DF inequalities \eqref{eq:tail2} and the ambient dimension (AD) inequality \eqref{eq:amb.tail} lies in their product factors: the former are with ${\rm e}^{(1+\alpha_1(\tau))\phi_\Omega}$ and the latter are with ${\rm dim}({\bf Y})$. Under the notations given in \eqref{eq:notationU}, we arrive at the following sufficient condition to guarantee that ${\rm e}^{(1+\alpha_1(\tau))\phi_\Omega}\leq {\rm dim}({\bf Y})$:
\begin{align}\label{eq:number}
I \leq \frac{\log {\rm dim}({\bf Y})}{\big[1+\alpha_1(\tau)\big]\cdot \big[ (\mu({\bf U})+1)^\tau-1       \big] } .
\end{align}
This condition suggests that the partition number $I$ should be in the order of $O(\log {\rm dim}({\bf Y}))$, and it meanwhile reflects that the DF inequalities \eqref{eq:tail2} are not suitable to the scenario of large quantities of summand matrices. However, there is a suboptimal method to overcome this limitation, that is, decreasing the magnitude of the random matrices ${\bf X}_k$ to generate a small $\mu({\bf U})$. We will demonstrate this strategy  with numerical experiments in Section \ref{sec:experiment}.

\end{remark}

{In addition, consider the function $g_4(\theta):= \theta^2 +\frac{1}{4}$. It is direct that $g_4(\theta) \geq \max\{\theta,\theta^2 \}$ for any $\theta>0$ and the curve of $g_4(\theta)$ is tangent to that of $\theta$ at the point $(\frac{1}{2},\frac{1}{2})$. By substituting $g_4(\theta)$ into Proposition \ref{prop:master2}, we then arrive at a Azuma-Hoeffding type tail inequalities:
\begin{theorem}\label{thm:tail5}
For any $t>0$, there holds that 
\begin{align}\label{eq:tail5}
\mathbb{P}\left\{\mu\left(\sum_{k=1}^K{\bf X}_k \right)\geq t \right\}  \leq  {\rm e}^{\frac{\phi_{\widetilde{\Omega}}}{4}}\cdot\exp\left\{ -\frac{t^2}{4\phi_{\widetilde{\Omega}}}\right\},
\end{align}
where $\phi_{\widetilde{\Omega}}:=\sum\limits_{i=1}^{\widetilde{I}}\big( \big[\mu\big(\widetilde{{\bf U}}_i\big)+1\big]^{|\widetilde{\Omega}_i|}-1\big)$ with $\widetilde{{\bf U}}_i = \mathop{\arg\max}\limits_{k\in\widetilde{\Omega}_i} \{\mu({\bf B}_k)  \}$.
\end{theorem}
Compared with Tropp's Azuma-Hoeffding type result \cite[Theorem 7.1]{tropp2012user}, our result has the following advantages:
\begin{enumerate}
\item it has no matrix dimension as a product factor;  
\item there is no restriction on the probability behavior of the random matrices ${\bf X}_k$; 
\item the matrix function $\mu(\cdot)$ can be set as many kinds of specific forms.
\end{enumerate}
Similar to \eqref{eq:number}, we can also obtain the following sufficient condition to guarantee that ${\rm e}^{\frac{\phi_{\widetilde{\Omega}}}{4}} \leq {\rm dim}({\bf Y})$:
\begin{align}\label{eq:number2}
\widetilde{I} \leq \frac{4\log {\rm dim}({\bf Y})}{\phi_{\widetilde{\Omega}}\cdot \big[ (\mu(\widetilde{{\bf U}})+1)^2-1       \big] } 
\end{align}
with $\widetilde{{\bf U}} := \mathop{\arg\max}\limits_{1\leq k\leq K} \{\mu({\bf B}_k)  \}$.

}

\subsection{An Empirical Method to Generate Fixed Matrices ${\bf B}_k$} \label{sec:selectb}

The obtained tail inequalities \eqref{eq:tail1} and \eqref{eq:tail2} rely on the existence of fixed matrices ${\bf B}_k$ that satisfy Condition \eqref{eq:cond.b}. In the following, we propose a {\it{constructive}} method to generate desired matrices ${\bf B}_k$ with high probability for the cases that 
(i) $\mu(\cdot)$ is the sum of the $j$ largest singular values for complex matrices, or (ii) it is the absolute value of the sum of the $j$ largest eigenvalues for Hermitian matrices.

 First, we present a sufficient condition for \eqref{eq:cond.b}.
\begin{proposition}\label{prop:validity}
 Let ${\bf X}\in \mathbb{M}$ be a random matrix. If there exists a fixed matrix ${\bf B}$ such that $\mathbb{E}\mu({\bf X})\leq \mu({\bf B})$, then it holds that 
\begin{equation*}
\mathbb{E}\,{\rm e}^{{\bf D}_\mu[\theta;{\bf X}]} \preceq {\rm e}^{{\bf D}_\mu[\theta;{\bf B}]}.
\end{equation*}
\end{proposition}
Hence, in order to guarantee the validity of Condition \eqref{eq:cond.b}, we only need to let the value of $\mu({\bf B}_k)$ be larger than or equal to the expectation $\mathbb{E}\mu({\bf X}_k)$, $1\leq k\leq K$. Then, the following theorem provides an empirical method to elavulate $\mu({\bf B}_k)$.

\begin{theorem}\label{thm:select}
Let ${\bf X}\in \mathbb{M}$ be a random matrix and ${\bf X}^{(1)},\cdots,{\bf X}^{(N)}\in\mathbb{M}$ be $N$ i.i.d.~observations of ${\bf X}$. For any $\gamma>0$, let the fixed matrix ${\bf B}_\gamma\in\mathbb{M}$ satisfy the relation
\begin{equation}\label{eq:bgamma}
\mu({\bf B}_\gamma) \geq \frac{1}{N}\left(\sum_{n=1}^N \mu({\bf X}^{(n)})\right)+\gamma\cdot \exp\left(\frac{1}{N}\sum_{n=1}^N \log \big(\mu( {\bf X}^{(n)})+1\big) \right).
\end{equation}
Then, with probability at least $1- \exp\Big(   \frac{-N (\log(1+\theta \gamma ) )^2}{2 \mathbb{E} (\log (\mu(\theta {\bf X})+1))^2     }  \Big)$, it holds
\begin{equation}\label{eq:cond.b2}
    \mathbb{E}\log \big(\mu(\theta {\bf X})+1\big)\leq  \log \big(\mu(\theta {\bf B}_\gamma)+1\big),\quad \theta>0.
\end{equation}
\end{theorem}
This theorem shows that 
if the fixed matrix ${\bf B}_\gamma$ satisfies the relation (\ref{eq:bgamma}), 
then the probability that \eqref{eq:cond.b} fails to hold will exponentially decay to {\it zero} as the observation number $N$ goes to {\it infinity}.


Finally, we explicitly demonstrate how to generate the fixed matrices ${\bf B}_k$ based on the estimated value of $\mu({\bf B}_k)$ for the following two cases of $\mu(\cdot)$.
\begin{enumerate}
\item Let ${\bf B}\in\mathbb{H}^{n\times n}$ and $\mu(\cdot)$ be the absolute value of the sum of the $j$ largest eigenvalues ($j\geq 1$). Denote $\lambda_1({\bf B})\geq \lambda_2({\bf B})\geq\cdots \geq\lambda_j({\bf B})$ as the $j$ largest eigenvalues of ${\bf B}$. For arbitrary $w_1\geq w_2\geq\cdots\geq w_j >0$ with $\sum_{i=1}^j  w_i  =1$, we can set $\lambda_i({\bf B}) = w_i \mu({\bf B})$ ($i=1,2,\cdots,j$). Then, the matrix ${\bf B}$ can be generated in the way of matrix eigenvalue decomposition:
\begin{equation*}
{\bf B}:= {\bf V}\cdot  \bm{\Lambda}[\lambda_1({\bf B}),\lambda_2({\bf B}),\cdots,\lambda_j({\bf B}),\underbrace{0,\cdots,0}_{n-j} ]  \cdot {\bf V}^*,
\end{equation*}
where ${\bf V}$ is an arbitrary $n\times n$ unitary matrix.

 \item Let ${\bf B}\in\mathbb{C}^{m\times n}$ and $\mu(\cdot)$ be the sum of the $j$ largest singular values ($j\geq 1$). Denote $\sigma_1({\bf B})\geq \sigma_2({\bf B})\geq\cdots \geq\sigma_j({\bf B})$ as the $j$ largest singular values of ${\bf B}$. Similarly, given $w_1\geq w_2\geq\cdots\geq w_j >0$ with $\sum_{i=1}^j  w_i  =1$, we can set $\sigma_i({\bf B}) = w_i \mu({\bf B})$ ($i=1,2,\cdots,j$). Then, the matrix ${\bf B}$ can be generated in the way of matrix singular value decomposition:
\begin{equation*}
{\bf B}:= {\bf U} \left[
\begin{array}{c c c c c c c}
  \sigma_1({\bf B}) & 0& \cdots  &  0 & 0  & \cdots & 0    \\
  0& \sigma_2({\bf B})  & \cdots  & 0 & 0 & \cdots & 0 \\
  \vdots& \vdots  &    \ddots  & \vdots &  \vdots& \ddots  & \vdots  \\
  0& 0  &    \cdots  & \sigma_j({\bf B}) & 0& \cdots  &0  \\
  0& 0  & \cdots  & 0 & 0 & \cdots & 0 \\
  \vdots& \vdots  &    \ddots  & \vdots &  \vdots& \ddots  & \vdots  \\
  0& 0  &    \cdots  & 0 & 0& \cdots  &0  \\
\end{array}
\right]_{m\times n}
  {\bf V}^*,
\end{equation*}
where ${\bf U}$ (resp. {\bf V}) can be an arbitrary $m\times m$ (resp. $n\times n$) unitary matrix.

\end{enumerate}


{\subsection{Dimension-free Tail inequalities for Matrix Random Series}

Matrix random series refers to sums of fixed matrices weighted by i.i.d. random variables, {\it i.e.,} it is of the form $\sum_{k=1}^K \xi_k {\bf A}_k$, where $\xi_1,\cdots,\xi_K$ are i.i.d. random variables and ${\bf A}_1,\cdots,{\bf A}_K\in\mathbb{C}^{m\times n}$ are fixed matrices. The study of matrix random series is motivated by applications of random matrices in 
neural networks \cite{zhao17theoretical}, kernel methods \cite{choromanski2016recycling}, deep learning \cite{cheng2015exploration} and optimization \cite{nemirovski2007sums,so2011moment,zhang2018matrix}, where the random matrices of interest can be equivalently expressed as matrix random series weighted by some specific random variables. One main research field on matrix random series is to explore their tail behaviors, and some tail results have been proposed. For example, Tropp \cite{tropp2012user} presented the tail inequalities for matrix Gaussian series and matrix Rademacher series, and his results can be directly generalized to the matrix sub-Gaussian series.\footnote{For convenience, the matrix random series weighted by Gaussian random variables is briefly named as the matrix Gaussian series, and this way of naming will be used in the whole paper if no confusion arises.} Zhang {\it et al.} \cite{zhang2018matrix} provided the tail inequalities for matrix infinitely-divisible series. There are two limitations in these works: 1) all of them are dependent on matrix dimension, and thus are unsuitable to the high-dimensional or infinite-dimensional scenario; and 2) they are only applicable to some specific distributions and thus are lack of generality.

The following dimension-free tail inequalities for matrix random series can be directly derived from Theorem \ref{thm:tail1} and Theorem \ref{thm:tail2}:

\begin{corollary}\label{cor:series}
Let ${\bf A}_1,\cdots,{\bf A}_K\in \mathbb{M}$ be fixed matrices and $\xi_1,\cdots\xi_K$ be independent random variables with $\max\limits_{1\leq k\leq K} \mathbb{E} |\xi_k | \leq c$. 
\begin{enumerate}
\item For any $t>0$, 
\begin{align}\label{eq:tail3}
\mathbb{P}\left\{\mu\left(\sum_{k=1}^K\xi_k{\bf A}_k \right)\geq t \right\}\leq& {\rm e}^{(1+\alpha_1(K))\cdot \psi}\cdot\exp\left(- \psi\cdot \Gamma\left(\frac{t}{K\psi}\right)\right) ,
\end{align}
where $\psi:=\left[c\mu\left({\bf V}\right)+1\right]^K-1$ with ${\bf V} = \mathop{\arg\max}\limits_{1\leq k\leq K} \{\mu({\bf A}_k)\}$.
\item For any $t>0$, 
\begin{align}\label{eq:tail4}
\mathbb{P}\left\{\mu\left(\sum_{k=1}^K\xi_k{\bf A}_k \right)\geq t \right\}\leq& {\rm e}^{\psi_\Omega(1+\alpha_1(\tau))}\cdot\exp\left(- \psi_\Omega\cdot \Gamma\left(\frac{t}{\tau\cdot \psi_\Omega}\right)\right),
\end{align}
where $\psi_\Omega:=\sum\limits_{i=1}^I\big( \big[c\mu\big({\bf V}_i\big)+1\big]^{|\Omega_i|}-1\big)$ with ${\bf V}_i = \mathop{\arg\max}\limits_{k\in\Omega_i} \{\mu({\bf A}_k)  \}$.

\end{enumerate}
\end{corollary}

Compared with the existing works \cite{tropp2012user,zhang2018matrix}, the above results have the following advantages: 1) they are independent of the matrix dimension, and thus are suitable to high-dimensional or infinite-dimensional scenario; 2) there is no requirement on the distributions except the bounded first-order moment, and thus they have better generality.}

\begin{remark}\label{rem:series}
The following is an application of the above results in optimization. The pioneering work \cite{nemirovski2007sums} and its follow-up \cite{so2011moment} have pointed out that whether there exist the efficiently computable solutions to some optimization problems ({\it e.g.}, chance constrained optimization problems and quadratic optimization problems with orthogonality constraints) can be reduced to a question about the tail behavior of matrix random series ({\it i.e.}, the upper bound of ${\rm Pr }\{ \|\sum_k \xi_k {\bf A}_k \|>t\}$), and the ``optimal" answer to this question will be provided by the resolution to Nemirovski's conjecture \cite{nemirovski2007sums}. The original version of Nemirovski's conjecture requires that the random variables $\xi_k$ should have {\it zero} mean and obey either distribution supported on $[-1,1]$ or Gaussian distribution with {\it unit} variance. 
Zhang {\it et al.} \cite{zhang2018matrix} extended Nemirovski's conjecture to the infinitely-divisible setting, where $\xi_k$ can be infinitely-divisible random variables. The resulted tail inequalities \eqref{eq:tail3} and \eqref{eq:tail4} actually suggest that Nemirovski's conjecture holds in a more general setting, where $\xi_k$ just have the bounded first-order moments. The detailed discussion is similar to that in \cite{zhang2018matrix}, so we omit it here.
 
\end{remark}

\subsection{Numerical Experiments}

At the end of this section, we conduct the experiments to empirically exam the validity of Theorem \ref{thm:select} and then to make a comparison between the AD tail inequality \eqref{eq:amb.tail} and the resulted dimension-free (DF) inequality \eqref{eq:tail2}.\footnote{In view of the comparability, we do not consider the ID inequality \eqref{eq:int.tail} in this experiment for two reasons: 1) its range of $t$ starts from $\sqrt{v} + L/3$ rather than the {\it origin}, and thus it cannot provides the comparative information when $t\in(0,\sqrt{v} + L/3)$; and 2) its product factor $4\cdot {\rm intdim}({\bf V})$ is likely to be much bigger than the factor ${\rm dim}({\bf Y})$ of the AD inequality \eqref{eq:amb.tail} in the experiments, and thus drawing curves of the ID inequality will decrease the readability of figures. }

\subsubsection{Examination of Theorem \ref{thm:select}} Consider the largest singular values of three types of random matrices whose entries obey the Gaussian distribution with {\it zero} mean and {\it unit} variance, the uniform distribution on $[-1,1]$ and the Rademacher distribution that takes $1$ or $-1$ with $1/2$ probability, respectively. The size of matrices is set as $50\times 10$ and let the constant $\theta=1$. {The expectation term $\mathbb{E}\log \big(\sigma_{\max}({\bf X})+1\big)$ is approximated by using the empirical term
\begin{equation*}
\frac{1}{3000}\sum_{n=1}^{3000} \log \big(\sigma_{\max}( {\bf X}^{(n)})+1\big),
\end{equation*}
where ${\bf X}^{(n)}$, $1\leq n\leq 3000$, are the independent observations of the random matrix ${\bf X}$. In this manner, the values of $\mathbb{E}\log \big(\sigma_{\max}({\bf X})+1\big)$ are approximately $2.3630$, $1.8681$ and $2.3408$ for the Gaussian random matrix, the uniform random matrix and the Rademacher random matrix, respectively.\footnote{{There have been many sophisticated results to prove the distributions of the largest singular values (or eigenvalues) of  specific random matrices, for example, the quadrant law for the singular values of Gaussian random matrices\cite{shen2001singular}, the semi-circle law for the eigenvalues of Gaussian orthogonal (or unitary) ensembles \cite{wigner1993characteristic} and Marchenko-Pastur law for the singular values of large rectangular random matrices \cite{marvcenko1967distribution}. However, these results are unsuitable (at least cannot be directly applied) to efficient computation of the expectation term $\mathbb{E}\log \big(\mu({\bf X})+1\big)$ for arbitrary applicable choices of the matrix function $\mu$. Therefore, we only adopt the empirical approximation of this term.}}}

\begin{figure}[htbp]
\centering
\includegraphics[height=8cm]{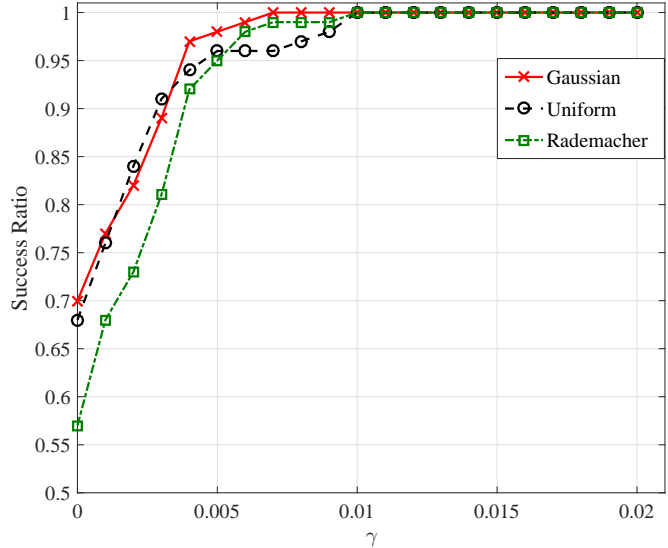}
\caption{Success Ratio of Inequality \eqref{eq:cond.b2} for Different Types of Random Matrices.}
\label{fig:condition17}
\end{figure}

 Given another set of independent observations ${\bf X}^{(1)},\cdots,{\bf X}^{(N)}$ of ${\bf X}$ with $N=100$, we compute ${\bf B}_\gamma$ according to the expression \eqref{eq:bgamma} and then exam the validity of the inequality \eqref{eq:cond.b2}. In Fig. \ref{fig:condition17}, we show the success ratios (out of 100 times repeated tests) of the inequality \eqref{eq:cond.b2} for different values of $\gamma\in(0,0.02]$. For these three kinds of random matrices, the success ratios (out of 100 times repeated tests) of the inequality \eqref{eq:cond.b2} all increase up to {\it one} as $\gamma$ becomes large, which supports the validity of Theorem \ref{thm:select}. 
 

\subsubsection{Examination of DF Tail Inequality}\label{sec:experiment} Let $\mu=\lambda_{\max}$ and {$\mathbb{M} = \mathbb{H}^{200\times 200}$.} Consider the random Hermitian matrices ${\bf X}_k = c \left( \frac{{\bf S}_k+{\bf S}_k^T}{2}\right)$, where $c$ is a positive constant to control the magnitude of ${\bf X}_k$ and the entries of ${\bf S}_k\in\mathbb{R}^{5\times 5}$ are all i.i.d. and obey the standard Gaussian distribution $\mathcal{N}(0,1)$. Thus $\mathbb{E}\mathrm{{\bf X}_k} = 0$, $1\leq k\leq K$. For each $k\in \{1,2 ,\cdots,K\}$, we take $1100$ observations $\widehat{{\bf S}}^{(i)}_k$ of ${\bf S}_k$ to generate the realizations {$\widehat{{\bf X}}^{(i)}_k = c\left(\frac{\widehat{{\bf S}}^{(i)}_k+(\widehat{{\bf S}}^{(i)}_k)^T}{2}\right)$}, $1\leq i\leq 1100$. To ensure that the probability $\mathbb{P}\big\{\big|\lambda_{\max}\big(\sum_k({\bf X}_k  \big)\big|\geq t \big\}$ will be strictly decreasing w.r.t. $t$, we alternatively consider the following probability expression $\mathbb{P}\big\{\big|\lambda_{\max}\big(\sum_k({\bf X}_k  \big) - \mathbb{E} \lambda_{\max}\big(\sum_k{\bf X}_k \big)\big|\geq t \big\}$, which is empirically computed by using the following function:
\begin{equation*}
h_{\rm TV}(t) = \frac{\big| \{ 1\leq i\leq 100 :  \big|\lambda_{\max}\big(\sum_{k=1}^K\widehat{{\bf X}}^{(i)}_k \big) - \frac{1}{1000}\sum\limits_{j=101}^{1100}\lambda_{\max}\big(\sum_{k=1}^K{\bf X}^{(j)}_k \big) \big|\geq t     \}\big|}{100},\quad t> 0.
\end{equation*}
In the AD inequality \eqref{eq:amb.tail}, the terms $v = \lambda_{\max}(\sum_k\mathbb{E} {\bf X}_k^2)$ and $L \geq \lambda_{\max}({\bf X}_k)$ are respectively approximated by using the empirical quantities 
\begin{equation*}
\widehat{v} = \lambda_{\max}\left(\sum_{k=1}^K \frac{1}{100} \sum_{i=1}^{100} \big(\widehat{{\bf X}}^{(i)}_k\big)^2\right),
\end{equation*}
and
\begin{equation*}
\widehat{L} = \mathop{\max_{1\leq k\leq K}}_{1\leq i\leq 100} \big\{ \lambda_{\max}(\widehat{\bf X}^{(i)}_k) \big\}.
\end{equation*}
Then, the right-hand sides of \eqref{eq:amb.tail} and \eqref{eq:tail2} can be respectively expressed as
\begin{align*}
h_{\rm AD} (t) := {\rm dim}({\bf X}_k)\cdot \exp \left(  \frac{-t^2/2}{\widehat{v}+\widehat{L}t/3} \right),&\quad t>0;\\
 h_{{\rm DF}}(t) := {\rm e}^{(1+\alpha_1(\tau))\cdot \phi_\Omega}\cdot\exp\left(-   \frac{t^2/2}{\tau^2 \phi_\Omega + \tau\cdot t/3}\right),&\quad t>0.
\end{align*}
The partition $\Omega$ of the index set $\{1,2,\cdots,K\}$ is designed according to the suggestion given in Remark \ref{rem:suggestion}. 

\begin{figure}[htbp]
\centering
\subfigure[\hbox{$K=5$, $c=1$}]{
\includegraphics[height=4.4cm]{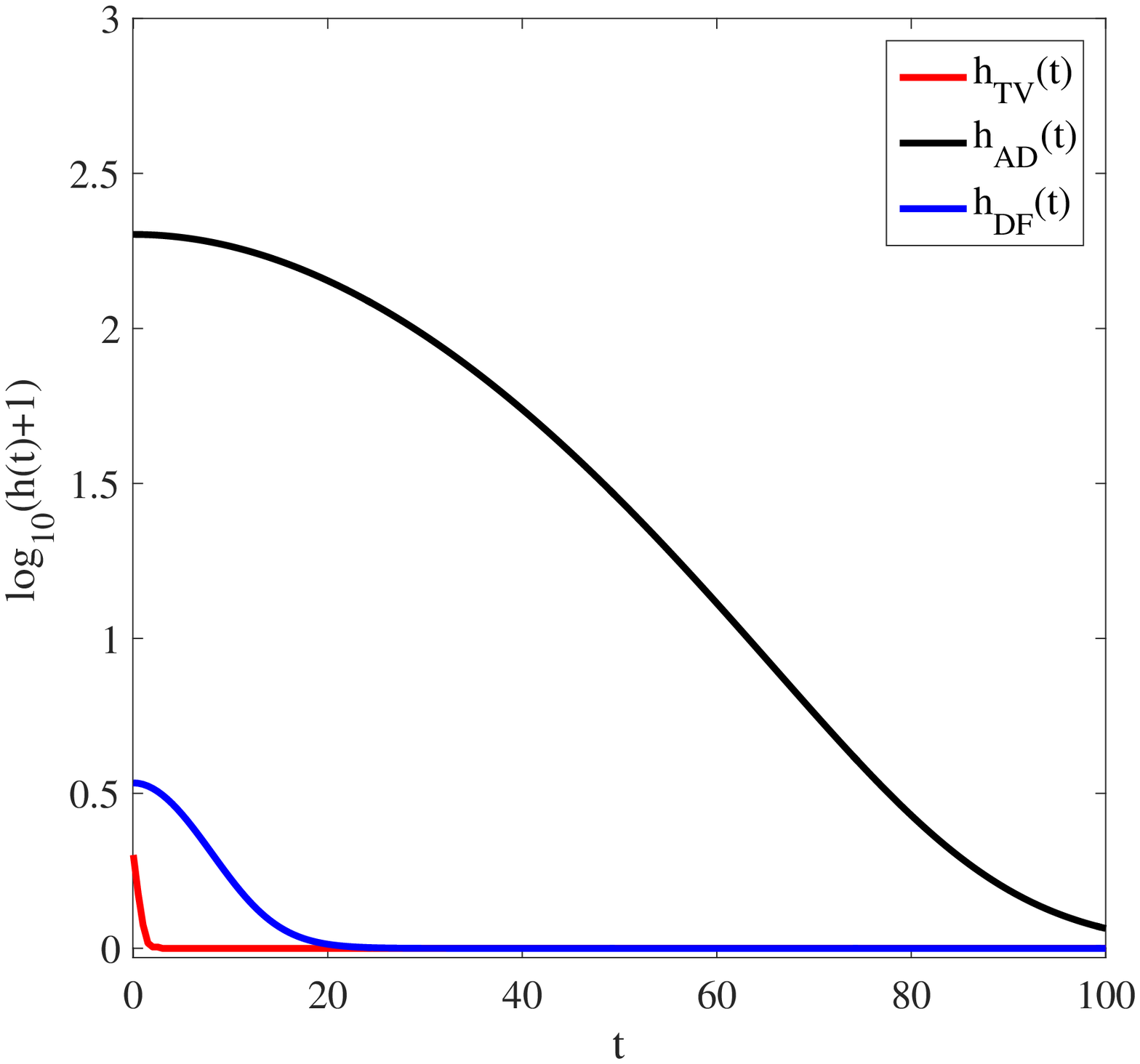} }
\subfigure[\hbox{$K=10$, $c=1$}]{
\includegraphics[height=4.4cm]{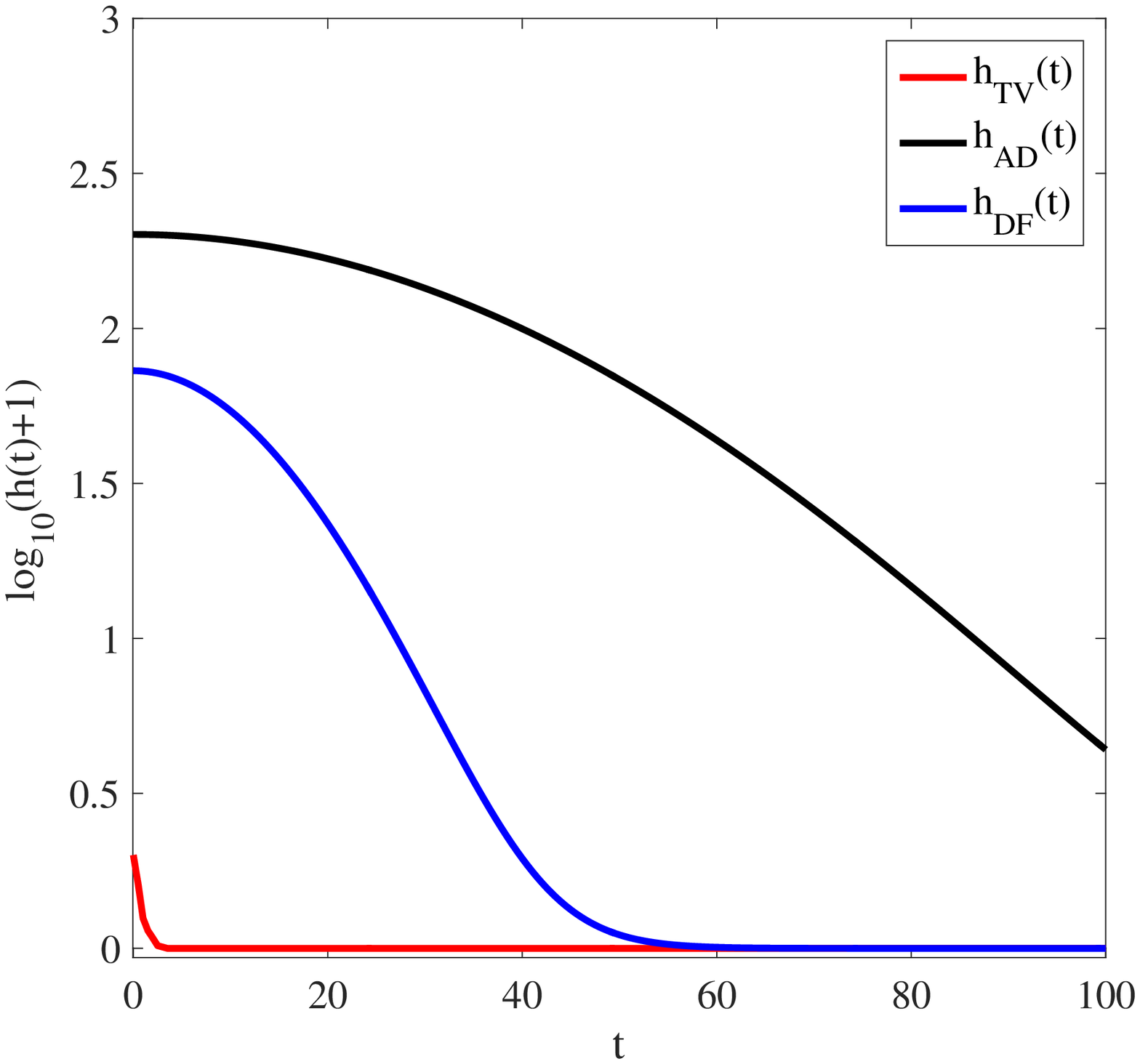}}
\subfigure[\hbox{$K=15$, $c=1$}]{
\includegraphics[height=4.4cm]{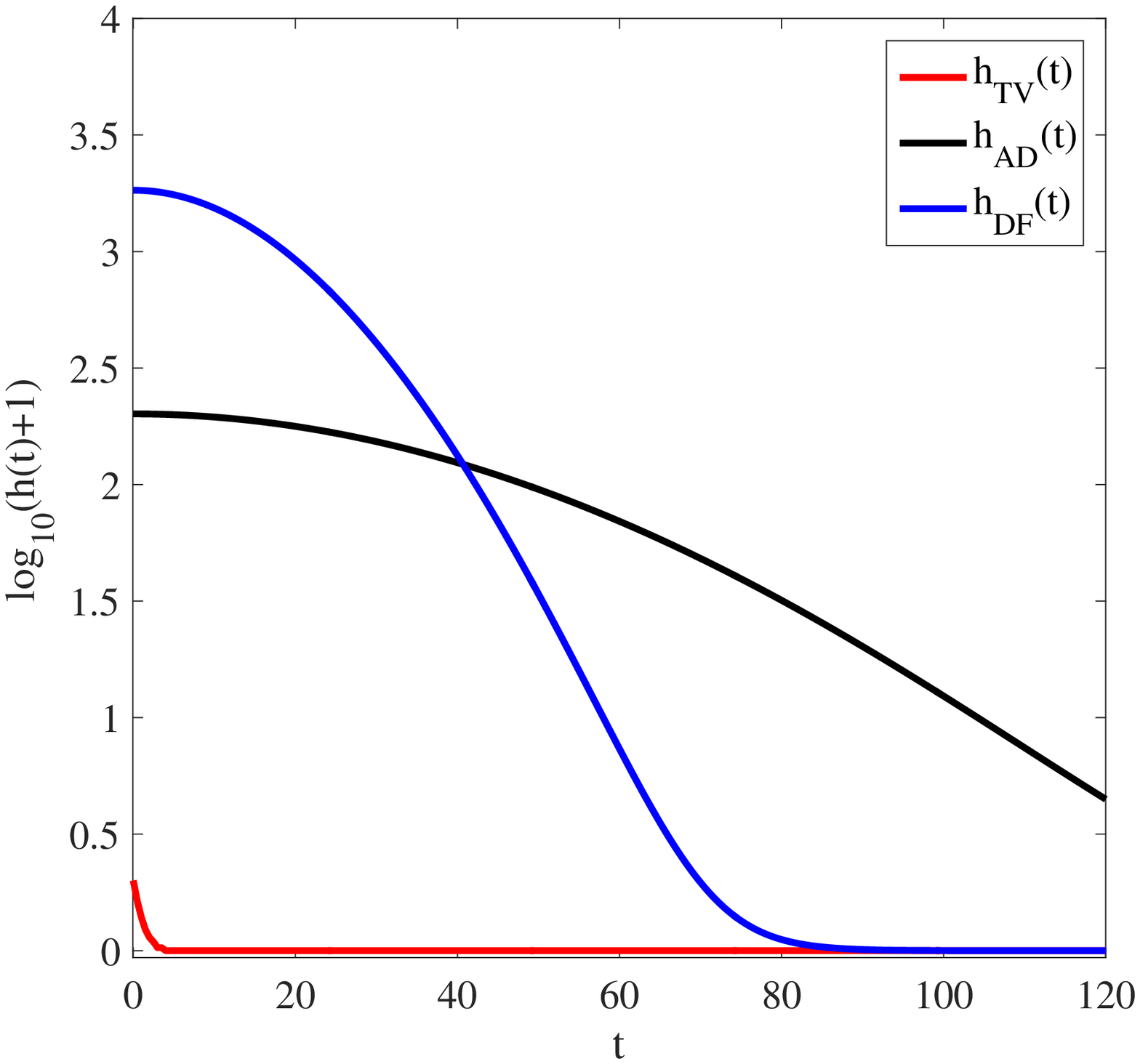}}
\subfigure[\hbox{$K=20$, $c=1$}]{
\includegraphics[height=4.4cm]{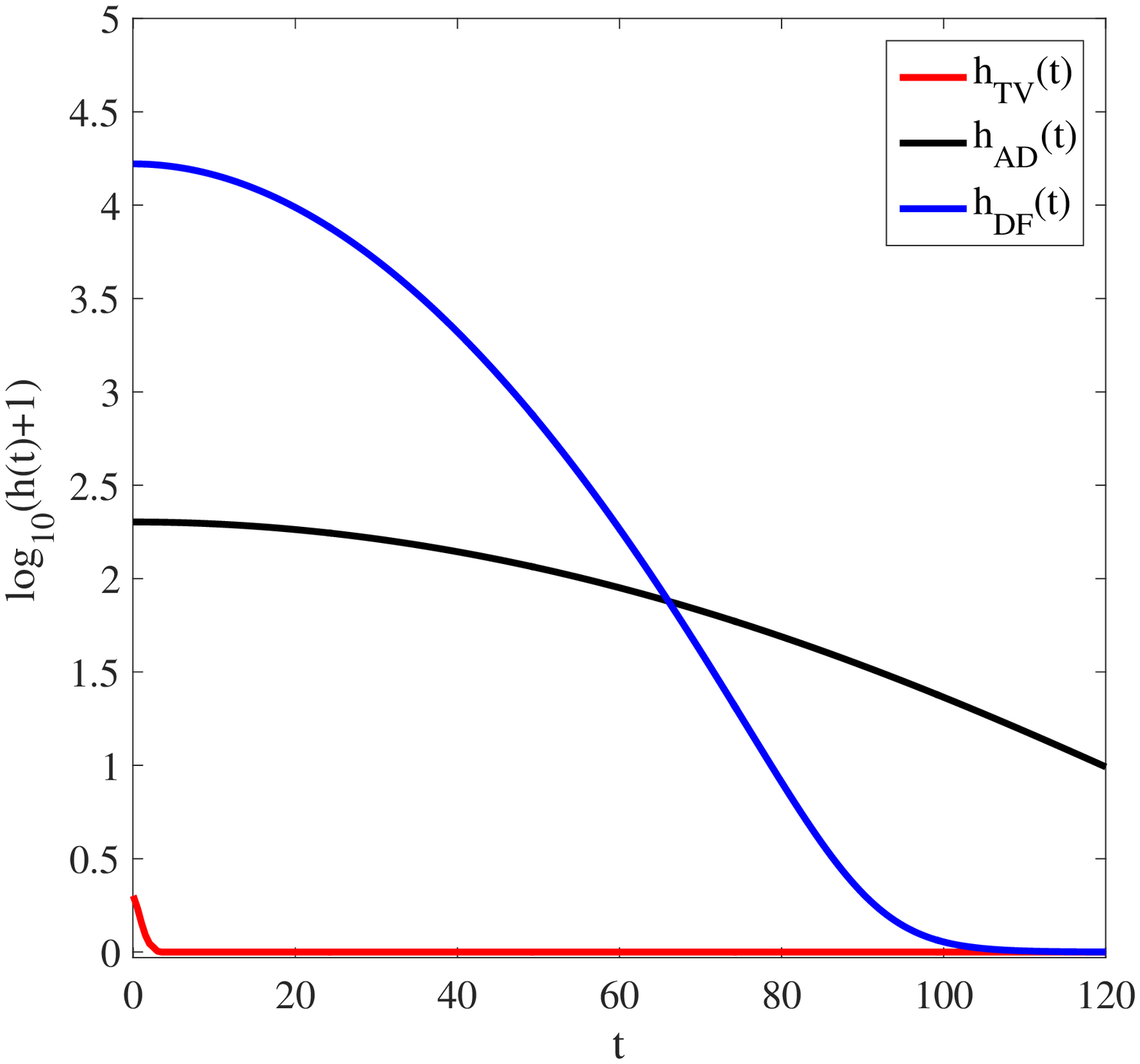}}
\subfigure[\hbox{$K=15$, $c=0.5$}]{
\includegraphics[height=4.4cm]{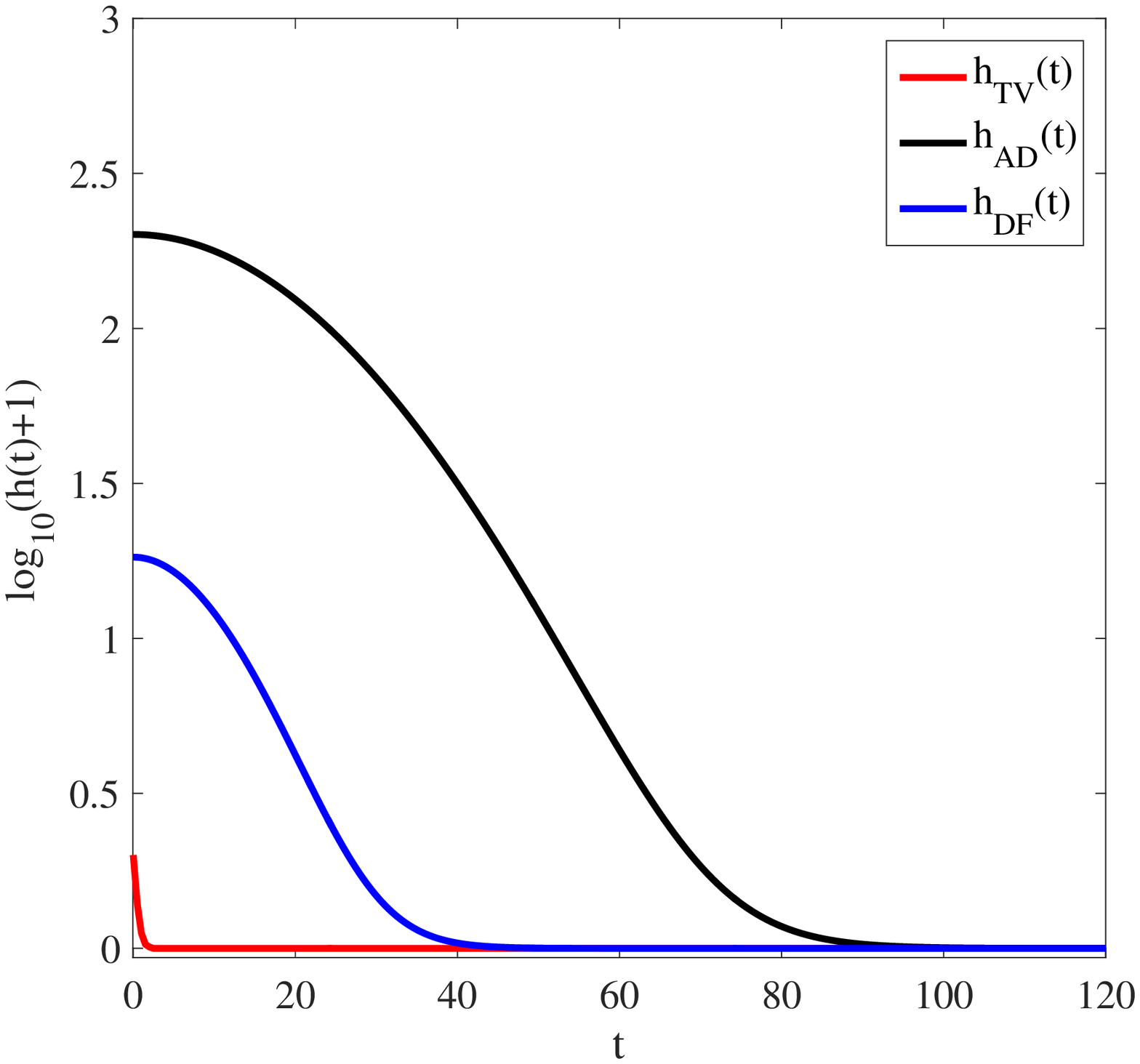} }
\subfigure[\hbox{$K=20$, $c=0.5$}]{
\includegraphics[height=4.4cm]{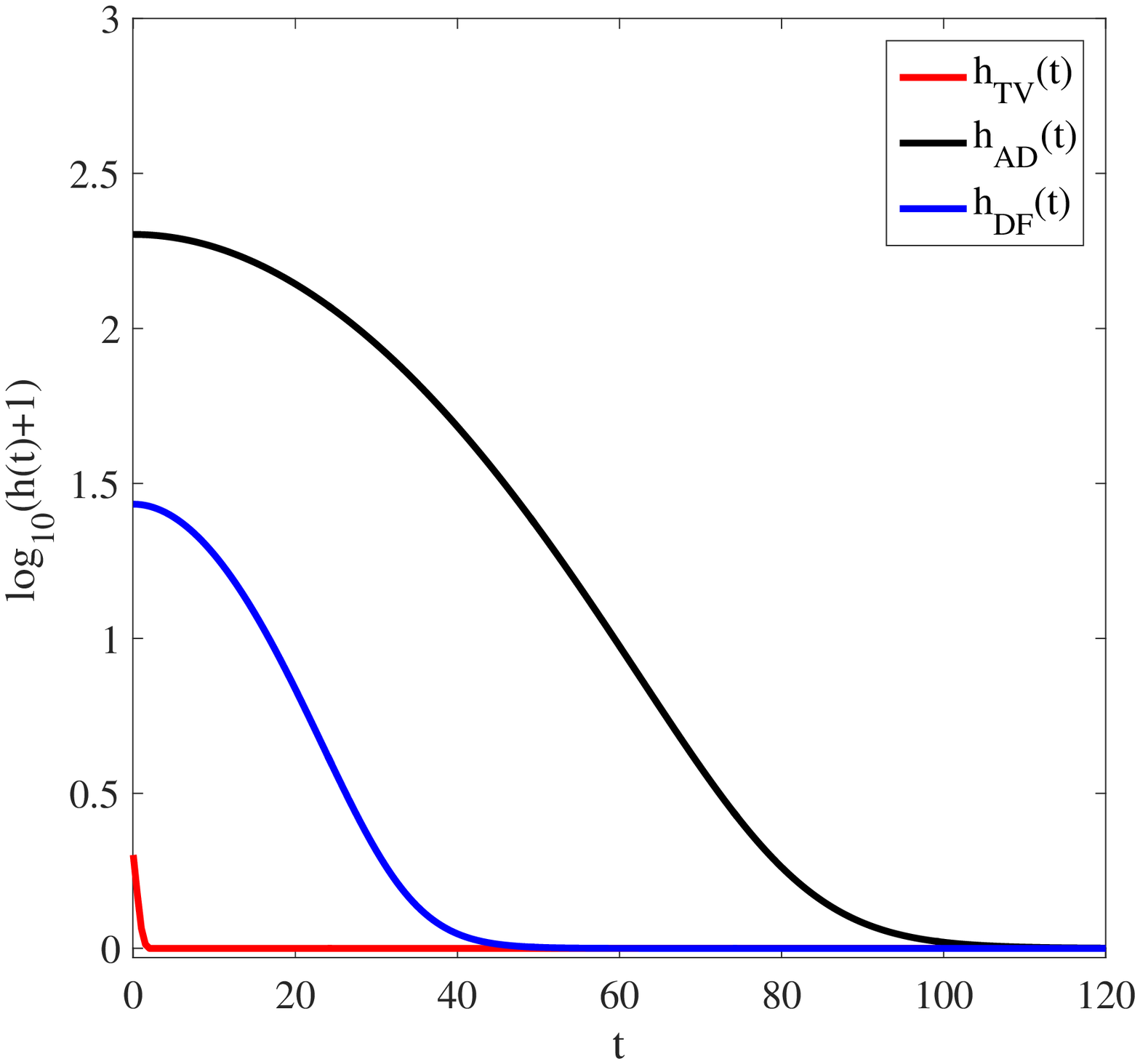}}

\caption{Numerical comparison among the AD inequality \eqref{eq:amb.tail} and the DF inequalities \eqref{eq:tail2}.}\label{fig:compare}
\end{figure}

As shown in Fig. \ref{fig:compare}, the DF inequality \eqref{eq:tail2} provides a precise description of the tail behavior of sums of random matrices when the summand number $K$ is small. However, if the value of $K$ increases, the value of $\phi_\Omega$ will become large and thus the upper bound of $h_{\rm TV}(t)$ provided by $h_{\rm DF}(t)$ turns out to be loose accordingly ({\it cf.} Fig. \ref{fig:compare}(c)-(d)). Following the statements in Remark \ref{rem:number}, we rescale the magnitude of the matrices ${\bf X}_k$ by setting $c<1$ to overcome this shortcoming ({\it cf.} Fig. \ref{fig:compare}(e)-(f)).

\section{Applications in Compressed Sensing}\label{sec:rip}

In this section, we show that the resulted tail inequalities can provide a simple proof of the restricted isometry property (RIP) of a measurement matrix that is expressed as the sum of random matrices without any assumption imposed on the distributions of matrix entries. We first give a brief introduction of RIP in compressed sensing, and then show the proof of RIP for sums of random matrices.


\subsection{Introduction of Restricted Isometric Property}

One core issue of compressed sensing is to recover a vector (or signal) ${\bf x}^\star \in\mathbb{C}^n$ by solving the underdetermined linear equation:
\begin{equation}\label{eq:cs.linear}
{\bf y} = {\bf P} {\bf x}^\star,
\end{equation}
where ${\bf y}\in\mathbb{C}^m$ ($m \ll n$) and ${\bf P} \in  \mathbb{C}^{m\times n}$ are called the measurement vector and the measurement (or sensing) matrix, respectively. This linear equation could have infinitely many solutions to this linear equation. Thus, by introducing an additional condition that ${\bf x}^\star$ is $s$-sparse, {\it i.e.,} $\| {\bf x}^\star\|_0 : = {\rm supp} ({\bf x}^\star) \leq s$,
this linear equation can be reformulated as an $\ell_0$-minimization problem:
\begin{equation}\label{eq:cs.l0min}
\min_{{\bf x}\in\mathbb{C}^N} \| {\bf x}\|_0\quad \mbox{subject to}  \quad {\bf P} {\bf x} = {\bf y}.
\end{equation}
One efficient way to solve this NP hard problem is to consider its $\ell_1$ convex relaxation {({\it cf.} \cite{candes2005decoding,candes2006stable,donoho2006compressed,ramirez2013l1})}:
\begin{equation}\label{eq:cs.l1min}
\min_{{\bf x}\in\mathbb{C}^n} \| {\bf x}\|_1\quad \mbox{subject to}  \quad {\bf P} {\bf x} = {\bf y},
\end{equation}
which can be solved with efficient convex optimization methods. Candes and Tao \cite{candes2005decoding,candes2006near} have proved that if the measurement matrix ${\bf P} \in  \mathbb{C}^{m\times n}$ satisfies the RIP, the recovery $\hat{\bf x}$ from the $\ell_1$-minimization \eqref{eq:cs.l1min} can approximate the true ${\bf x}^\star$ well. Hence, the RIP plays an essential role in compressed sensing.

\begin{definition}[Restricted Isometry Property]\label{def:rip1}
Given a matrix ${\bf P}\in\mathbb{C}^{m\times n}$, for any $0\leq s\leq n$, the restricted isometry constant of ${\bf P}$ of order $s$  is defined as the smallest number $\delta_s:=\delta_s({\bf P})$ such that
\begin{equation}\label{eq:rip1}
(1-\delta_s )\| {\bf x}\|_2 \leq \|{\bf P} {\bf x}  \|_2 \leq (1+\delta_s) \| {\bf x}\|_2,
\end{equation}
for all $s$-sparse ${\bf x}\in\mathbb{C}^n$. Let $\delta\in(0,1)$, the matrix ${\bf P}$ is said to satisfy the restricted isometry property (RIP) of order $s$ with parameter $\delta$, shortly, ${\rm RIP}_s(\delta)$, if $0\leq \delta_s({\bf P}) <\delta$.
\end{definition}

In the literature, many types of measurement matrices have been proven to satisfy the RIP condition with high probability, {\it e.g.}, random Gaussian or Bernoulli matrices ({\it cf.} \cite{candes2006near,mendelson2008uniform}); the structured matrices with Gaussian or Bernoulli entries ({\it cf.} \cite{haupt2010toeplitz,rauhut2009circulant}); {and the matrix infinitely-divisible series \cite{zhang2018matrix}.} Different from these existing works that specify the probability distributions of entries of measurement matrices, we will show that if the measurement matrix ${\bf P}\in\mathbb{C}^{m\times n}$ can be expressed as the sum of random matrices ${\bf P}_1,\cdots, {\bf P}_K$ that satisfy a mild condition \eqref{eq:smallest.sigma}, its RIP still holds with a high probability. To achieve this proof, we will borrow the idea of the work \cite{baraniuk2008simple}, where by introducing an alternative definition of RIP, Baraniuk {\it et al.} proved the RIP of a measurement matrix under the assumption that it satisfies a concentration inequality \cite[Eq. (4.3)]{baraniuk2008simple}. We will directly use the resulted tail inequalities to prove the RIP of ${\bf P}=\sum_{k=1}^K {\bf P}_k$ without imposing any distribution assumption on ${\bf P}$.  

\subsection{RIP of Sums of Random Matrices}

Given a matrix ${\bf P}\in\mathbb{C}^{m\times n}$ and any set $\mathcal{I}\subseteq \{1,2,\cdots,n\}$ of column indices, denote by $[{\bf P}]_\mathcal{I}$ the $m\times |\mathcal{I}| $ matrix composed of these columns, where $|\mathcal{I}|$ stands for the cardinality of the set $\mathcal{I}$. Similarly, for a vector ${\bf x}\in\mathbb{C}^n$, we denote ${\bf x}_\mathcal{I} $ as the $|\mathcal{I}|$-dimensional vector obtained by retaining only the entries in ${\bf x}$ corresponding to the column indices in $\mathcal{I}$. Alternatively, under these notations, Baraniuk {\it et al.} \cite{baraniuk2008simple} introduced another version of the RIP definition: a matrix ${\bf P}\in\mathbb{C}^{m\times n}$ is said to satisfy the ${\rm RIP}_s(\delta)$ if there exists a $\delta_s\in(0,1)$ such that
\begin{equation}\label{eq:rip2}
(1-\delta_s) \| {\bf x}_\mathcal{I}\|_2 \leq \|[{\bf P}]_\mathcal{I} {\bf x}_\mathcal{I}  \|_2 \leq (1+\delta_s) \| {\bf x}_\mathcal{I}\|_2,
\end{equation}
holds for all sets $\mathcal{I}$ with $|\mathcal{I}| \leq s$. As shown in \eqref{eq:rip2}, the ${\rm RIP}_s(\delta)$ requires that all singular values of $[{\bf P}]_\mathcal{I}$ lie in the interval $[1-\delta_s,1+\delta_s]$ for any $\mathcal{I}\subset \{1,\cdots,n\}$ with $|\mathcal{I}| \leq s$.

\begin{theorem}\label{thm:rip}
Let ${\bf P}_1,\cdots,{\bf P}_K \in \mathbb{C}^{m\times n}$ be random matrices and ${\bf P} =\sum_{k=1}^K  {\bf P}_k$. Given a number $s<n$, assume that  
\begin{equation}\label{eq:smallest.sigma}
\sigma_{\min} ([{\bf P}]_\mathcal{I})  \geq \frac{\sigma_{\min} ([{\bf P}_1]_\mathcal{I})+\cdots + \sigma_{\min} ([{\bf P}_K]_\mathcal{I}) }{K^2}
\end{equation}
holds for any $\mathcal{I} \subset \{1,2,\cdots,n\}$ with $|\mathcal{I}| =s$, where $[{\bf P}]_{\mathcal{I}}$ stands for the $m\times |\mathcal{I}| $ matrix composed of the columns taken from the matrix ${\bf P}$ w.r.t. the index set $\mathcal{I}$. Let $\{{\bf A}_{1},\cdots,{\bf A}_{K}\}$ and $\{{\bf B}_{1},\cdots,{\bf B}_{K}\}$ be the two fixed matrix sequences such that for any $\mathcal{I} \subset \{1,2,\cdots,n\}$,
\begin{align}\label{eq:condition.bc}
\mathbb{E}\,{\rm e}^{{\bf D}_{\sigma_{\max}}[\theta;[{\bf P}_k]_\mathcal{I}]} \preceq {\rm e}^{{\bf D}_{\sigma_{\max}}[\theta;[{\bf A}_{k}]_\mathcal{I}]}\;\;\mbox{and}\;\;  \mathbb{E}\,{\rm e}^{{\bf D}_{\sigma_{\max}}[\theta;[{\bf P}_k]_\mathcal{I}^{\dag}]} \preceq {\rm e}^{{\bf D}_{\sigma_{\max}}[\theta;[{\bf B}_{k}]_\mathcal{I}]}, \quad 1\leq k\leq K,
\end{align}
where the superscript $\dag$ stands for the Moore-Penrose inverse. Let $\Omega = \{\Omega_1,\cdots,\Omega_I\}$ be a partition of the index set $\{1,\cdots, K\}$ with $\bigcup_{i=1}^I \Omega_i=\{1,\cdots, K\}$ and $\tau: = \max\limits_{1\leq i\leq I}\{ |\Omega_i|\}$. {For any $1\leq i\leq I$, let 
\begin{align*}
u_i :=& \mathop{\max_{ k\in\Omega_i  }}_{ \mathcal{I} \subset \{1,2,\cdots,n\}} \Big\{
 \sigma_{\max}([{\bf A}_{k}]_\mathcal{I} )\Big\};\nonumber\\
v_i :=& \mathop{\max_{ k\in\Omega_i  }}_{ \mathcal{I} \subset \{1,2,\cdots,n\}} \Big\{
 \sigma_{\max}([{\bf B}_{k}]_\mathcal{I} )\Big\},
\end{align*}
and denote 
\begin{align*}
\bar{\phi}_\Omega = \max\left\{ \sum_{i=1}^I \Big[\big(u_i+1\big)^{|\Omega_i|} -1\Big],
\sum_{i=1}^I \Big[\big(v_i+1\big)^{|\Omega_i|} -1\Big]    \right\}.
\end{align*}}
For any $\delta\in(0,1)$, if there exists two positive constants $c_1$ and $c_2$ such that 
\begin{equation}\label{eq:rip.cond1}
s \leq \frac{ c_1m    }{ \log({\rm e} n/s) },
\end{equation}
and
\begin{equation}\label{eq:rip.cond2}
c_2\leq \frac{\bar{\phi}_\Omega}{m}\cdot \Gamma\left(\frac{t}{\tau\cdot \bar{\phi}_\Omega}\right) -c_1,
\end{equation}
then the ${\rm RIP}_s(\delta)$ \eqref{eq:rip1} holds for the random matrix ${\bf P}$ with probability at least $1-2{\rm e}^{\bar{\phi}_\Omega(1+\alpha_1(\tau))}\cdot {\rm e}^{-c_2 m}$.
\end{theorem}

Note that the tail inequalities \eqref{eq:tail2} can also lead to the similar RIP results, and we omit them here. The validity of this theorem is determined by the following factors: 1) the validity of Condition \eqref{eq:smallest.sigma}; 2) the existence of ${\bf A}_k$ and ${\bf B}_k$; and 3) Conditions \eqref{eq:rip.cond1} and \eqref{eq:rip.cond2} that can be satisfied by selecting sufficiently small $c_1>0$. Subsequently, we will give a detailed discussion on these factors.

\begin{figure}[htbp]
\centering
\includegraphics[height=8cm]{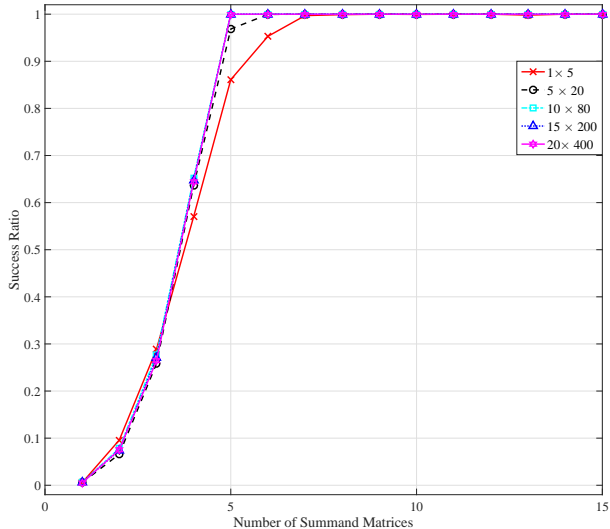}
\caption{Success Ratio of Condition \eqref{eq:smallest.sigma} for Different Matrix Sizes.}
\label{fig:condition51}
\end{figure}

\begin{remark}\label{rem:rip.condition}

{To examine the validity of condition \eqref{eq:smallest.sigma}, we alternatively consider whether it holds for any ${\bf A}_1,\cdots,{\bf A}_K \in\mathbb{C}^{m\times n}$ that
\begin{equation*}
\sigma_{\min} ({\bf A})  \geq \frac{\sigma_{\min} ({\bf A}_1)+\cdots + \sigma_{\min} ({\bf A}_K) }{K^2}\quad\mbox{with}\quad {\bf A} = \sum_{k=1}^K {\bf A}_k.
\end{equation*}
This inequality, roughly speaking, requires that the summands ${\bf A}_1,{\bf A}_2,\cdots,{\bf A}_K$ should not make ${\bf A}$ have {\it zero} singular values.} Here, we design an experiment to empirically verify the validity of {this inequality}. 
Let the number $K$ of the summand matrixes ${\bf A}_k$  evaluate from the set $\{2,5,10,15,20,25,30\}$ and the matrix sizes $m\times n$ of ${\bf A}$ are set as $1\times 5$, $5\times 20$, $10\times 80$, $15\times 200$ and $20\times 400$, respectively. Let the entries of ${\bf A}_k$ obey the Gaussian distribution with {\it zero} mean and {\it unit} variance. For each experimental setting, repeat $2000$ times and the success ratio of Condition \eqref{eq:smallest.sigma} is shown in Fig. \ref{fig:condition51}. We find that the success ratio can mostly reach $1$ when $K$ is larger than $5$, which implies that Condition \eqref{eq:smallest.sigma} can be easily satisfied when the summand matrices $\mathbf{A}_k$ are not too few. Especially, the high matrix dimension is beneficial to the success ratio {as well}. 

\end{remark}

\begin{remark}
To guarantee the existence of ${\bf A}_k$ and ${\bf B}_k$, we first need to consider the validity of Condition \eqref{eq:condition.bc}, which aims to control the random behavior of the terms $[{\bf P}_k]_\mathcal{I}$ and $[{\bf P}_k]_\mathcal{I}^{\dag}$. As addressed in Proposition \ref{prop:validity}, to safisfy this condition, the fixed matrices ${\bf A}_{k}$ and ${\bf B}_{k}$ should guarantee that the inequalities $\sigma_{\max}([{\bf A}_{k}]_\mathcal{I}) \geq \mathbb{E}\sigma_{\max}( [{\bf P}_k]_\mathcal{I} )$ and $\sigma_{\max}([{\bf B}_{k}]_\mathcal{I}) \geq \mathbb{E} \sigma_{\max}( [{\bf P}_k]_\mathcal{I}^{\dag} )$ hold for any $\mathcal{I}\subset\{1,\cdots,n\}$ with $|\mathcal{I}|=s$. Next, we will show how to construct such fixed matrices ${\bf A}_{k}$ and ${\bf B}_{k}$. {For any $k\in\{1,2,\cdots,K\}$, denote
\begin{align*}
a_k :=&\max_{ \mathcal{I}  }\Big\{ \mathbb{E}\sigma_{\max}( [{\bf P}_k]_\mathcal{I} ) \Big\};\\
b_k :=&\max_{ \mathcal{I}  } \Big\{ \mathbb{E} \sigma_{\max}( [{\bf P}_k]_\mathcal{I}^{\dag} )\Big\},
\end{align*}
and then let the fixed matrices ${\bf A}_k$ and ${\bf B}_k$ have the following forms:
\begin{equation*}
{\bf A}_k:= \left[
\begin{array}{c c c c c c c c c c c c c  c c }
 a_k &0&0&0     &  a_k &0&0&0  &   a_k & 0 & 0 & 0 & a_k & 0 &\cdots  \\ 
0&a_k&0&0    &    0&a_k&0&0 &     0&  a_k & 0 & 0 & 0&  a_k &\cdots \\ 
\vdots&\vdots& \ddots & \vdots   &  \vdots&\vdots& \ddots & \vdots &  \vdots&  \vdots&  \ddots & \vdots &  \vdots&  \vdots& \vdots\\
0&0&0&  a_k   &  0&0&0&  a_k &    0& 0&   0&  a_k &  0&  0& \cdots
\end{array}\right]_{m\times n},
\end{equation*}
and 
\begin{equation*}
{\bf B}_k:=\left[
\begin{array}{c c c c c c c c c c c c c  c c }
 b_k &0&0&0     &  b_k &0&0&0  &   b_k & 0 & 0 & 0 & b_k & 0 &\cdots  \\ 
0&b_k&0&0    &    0&b_k&0&0 &     0&  b_k & 0 & 0 & 0&  b_k &\cdots \\ 
\vdots&\vdots& \ddots & \vdots   &  \vdots&\vdots& \ddots & \vdots &  \vdots&  \vdots&  \ddots & \vdots &  \vdots&  \vdots& \vdots\\
0&0&0&  b_k   &  0&0&0&  b_k &    0& 0&   0&  b_k &  0&  0& \cdots
\end{array}\right]_{m\times n}.
\end{equation*}
Taking the matrix ${\bf A}_k$ as example, we first consider some special cases:
\begin{itemize}
\item If the index set $\mathcal{I}$ makes the $s$ column vectors selected from ${\bf A}_k$ differ from each other, the matrix product $[{\bf A}_k]_\mathcal{I}^T \cdot  [{\bf A}_k]_\mathcal{I}$ is a diagonal matrix with the identical entries $a^2_k$ and thus $\sigma_{\max}([{\bf A}_k]_\mathcal{I})=a_k$.

\item In the case that $\frac{n}{m} \geq s$, if the index set $\mathcal{I}$ takes $s$ identical column vectors from ${\bf A}_k$ to form $[{\bf A}_k]_\mathcal{I}$, the matrix product $[{\bf A}_k]_\mathcal{I}^T \cdot  [{\bf A}_k]_\mathcal{I}$ is a diagonal matrix with only one non-zero entry $s\cdot a^2_k$ and thus $\sigma_{\max}([{\bf A}_k]_\mathcal{I})=\sqrt{s}\cdot a_k$; 

\item In the case that $\frac{n}{m} < s$, if the index set $\mathcal{I}$ selects $\lceil \frac{n}{m}\big \rceil$ identical and $s - \lceil \frac{n}{m}\big \rceil $ different column vectors from ${\bf A}_k$ to form $[{\bf A}_k]_\mathcal{I}$, the matrix product $[{\bf A}_k]_\mathcal{I}^T \cdot  [{\bf A}_k]_\mathcal{I}$ is a diagonal matrix with $(s+1-\lceil \frac{n}{m}\big \rceil)$ non-zero entries: one is $\lceil \frac{n}{m}\big \rceil\cdot a^2_k$ and the others are $a^2_k$. Thus, we have $\sigma_{\max}([{\bf A}_k]_\mathcal{I})= \sqrt{ \big\lceil \frac{n}{m}\big \rceil}\cdot a_k$, where $\lceil \cdot \rceil$ stands for the ceiling function. 

\end{itemize}
Without loss of generality, we then have, for any $\mathcal{I} \subset \{1,2,\cdots, n\}$ with $|\mathcal{I}| =s$, 
\begin{align*}
a_k& \leq \sigma_{\max}([{\bf A}_k]_\mathcal{I}) \leq 
\left\{
\begin{array}{l l }
 \sqrt{ s } \cdot a_k,&   \mbox{if  $\frac{n}{m} > s$; }   \\
   \sqrt{ \big\lceil \frac{n}{m}\big \rceil} \cdot a_k, &    \mbox{otherwise,}   
\end{array}
\right.\\
b_k& \leq \sigma_{\max}([{\bf B}_k]_\mathcal{I}) \leq 
\left\{
\begin{array}{l l }
 \sqrt{ s } \cdot b_k,&   \mbox{if  $\frac{n}{m} >s$; }   \\
   \sqrt{ \big\lceil \frac{n}{m}\big \rceil} \cdot b_k, &    \mbox{otherwise.}   
\end{array}
\right.
\end{align*}}
In this manner, the resulted matrices ${\bf A}_k$ and ${\bf B}_k$ satisfy Condition \eqref{eq:condition.bc} for any $k\in\{1,2,\cdots, K\}$.

\end{remark}


{\section{Applications in Stochastic Processes}\label{sec:process}

The supremum of stochastic processes has long been an important issue in the field of probability theory. Hsu {\it et al.} \cite{hsu2011dimension} embedded a stochastic process into an infinite-dimensional diagonal random matrix, and then used the tail inequalities of random matrices to solve this issue. Here, we borrow this embedding idea to analyze the supremum of a stochastic process by applying the resulted DF tail inequalities.

Let $\{X_1,X_2,X_3,\cdots\}$ be a stochastic process with a constant $\beta$ such that $\mathbb{E} |X_i| \leq \beta$ holds for any $i=1,2,\cdots$. Let ${\bf X} := \bm{\Lambda}[X_1,X_2,X_3,\cdots]$ be an infinite-dimensional diagonal random matrix. Letting $\mu = \sigma_{\rm max}$, then it follows from Theorem \ref{thm:tail1} and {$\phi \leq \beta$} that 
\begin{align*}
\mathbb{P}\left\{\sup_{i>0} |X_i|\geq t \right\} =& \mathbb{P}\left\{\sigma_{\rm max}({\bf X} )\geq t \right\}\nonumber\\
\leq & \left\{
\begin{array}{cc}
  {\rm e}^{(2\log2-1)\beta }\cdot {\rm e}^{\frac{-t^2}{ 4 \beta }} , &  \mbox{if $t< 3\beta $;}    \\
   {\rm e}^{(2\log2-1)\beta }\cdot {\rm e}^{\frac{-3t}{ 4 }}, &    \mbox{if $t\geq 3\beta $.}   
\end{array}
\right.
\end{align*}
Alternatively, the above expression can be equivalently rewritten as
\begin{align}\label{eq:rp.supremum}
 \left\{
\begin{array}{ll}
 \mathbb{P}\left\{\sup\limits_{i>0} |X_i|\geq \sqrt{ 4\beta \big(\epsilon +(2\log2-1)\cdot\beta\big)   }   \right\} \leq {\rm e}^{-\epsilon} , &  \mbox{if $t< 3\beta $;}    \\
  \mathbb{P}\left\{\sup\limits_{i>0} |X_i|\geq  \frac{4}{3}\big(\epsilon +(2\log2-1)\cdot\beta\big)   \right\} \leq  {\rm e}^{-\epsilon}, &    \mbox{if $t\geq 3\beta $,}   
\end{array}
\right.
\end{align}
Compared with the existing work \cite{hsu2011dimension}, this result is independent of matrix dimension and applicable to the various kinds of stochastic processes as long as the expectation $\mathbb{E} |X_i|$ ($\forall i>0$) has a unified upper bound.

\section{Applications in Matrix Approximation via Random Sampling}\label{sec:approximation}

Matrix approximation via random sampling aims to estimate a complicated objective matrix by constructing some structural random matrices whose expectations are identical with the objective matrix, and has been widely used in many practical applications of linear algebra and machine learning, {\it e.g.,} matrix random sparsification \cite{drineas2011note}, randomized matrix multiplication \cite{drineas2006fast,magen2011low} and random feature \cite{rahimi2008random,lopez2014randomized}. 

Tail bounds of random matrices provide analytical benchmarks to the approximation quality of the matrix-approximation strategy. Tropp \cite{tropp2015introduction} applied the dimension-dependent expectation bounds for sums of random matrices to provide a comprehensive analysis on these applications. Hsu {\it et al.} \cite{hsu2011dimension} obtained the upper bound of probability that the discrepancy between the estimator and the objective matrix is large in the randomized matrix multiplication. Nevertheless, their results are dependent on the matrix dimension. Here, we will explore the properties of matrix approximation via random sampling from the dimension-free viewpoint.

\subsection{Dimension-free Expectation Bounds}
 
The following lemma is a part of the proof of Proposition \ref{prop:diagonal.sum}:

\begin{lemma}\label{lem:exp.bound}
For any $\theta>0$, it holds that
\begin{equation}\label{eq:exp.bound00}
\mathbb{E}{\rm e}^{\mu\left(\sum_{k=1}^K\theta {\bf X}_k\right)} \leq {\rm e}^{g(\theta,\tau)\cdot \phi_\Omega}.
\end{equation}
\end{lemma}

Consider the function 
\begin{equation}\label{eq:g3}
g_2(\theta;c) := \frac{3\theta^2}{6-2c\theta} +\alpha_2(c),\quad 0<\theta<\frac{3}{c},\;c>0
\end{equation}
with
\begin{equation}\label{eq:alpha3}
\alpha_2(c) = \frac{3[(c+3) - \sqrt{6c+9}] }{c^2}\geq 0.
\end{equation}
It holds that $g_2(\theta;c) \geq  \max\{\theta,\theta^2\}$ for any $0<\theta<\frac{3}{c}$ with $c > 0$. The curve of $g_2(\theta;c)$ is tangent to that of $\theta$ at the point $\Big(\frac{(6c+9)-3\sqrt{6c+9}}{2c^2+3c},\frac{(6c+9)-3\sqrt{6c+9}}{2c^2+3c}\Big)$, and is illustrated in Fig. \ref{fig:g3} for different values of $c$.

\begin{figure}[htbp]
\centering
\subfigure[\hbox{The curves of $g_2(\theta;c)$ and $\max\{ \theta ,\theta^2\}$.}]{
\includegraphics[height=6cm]{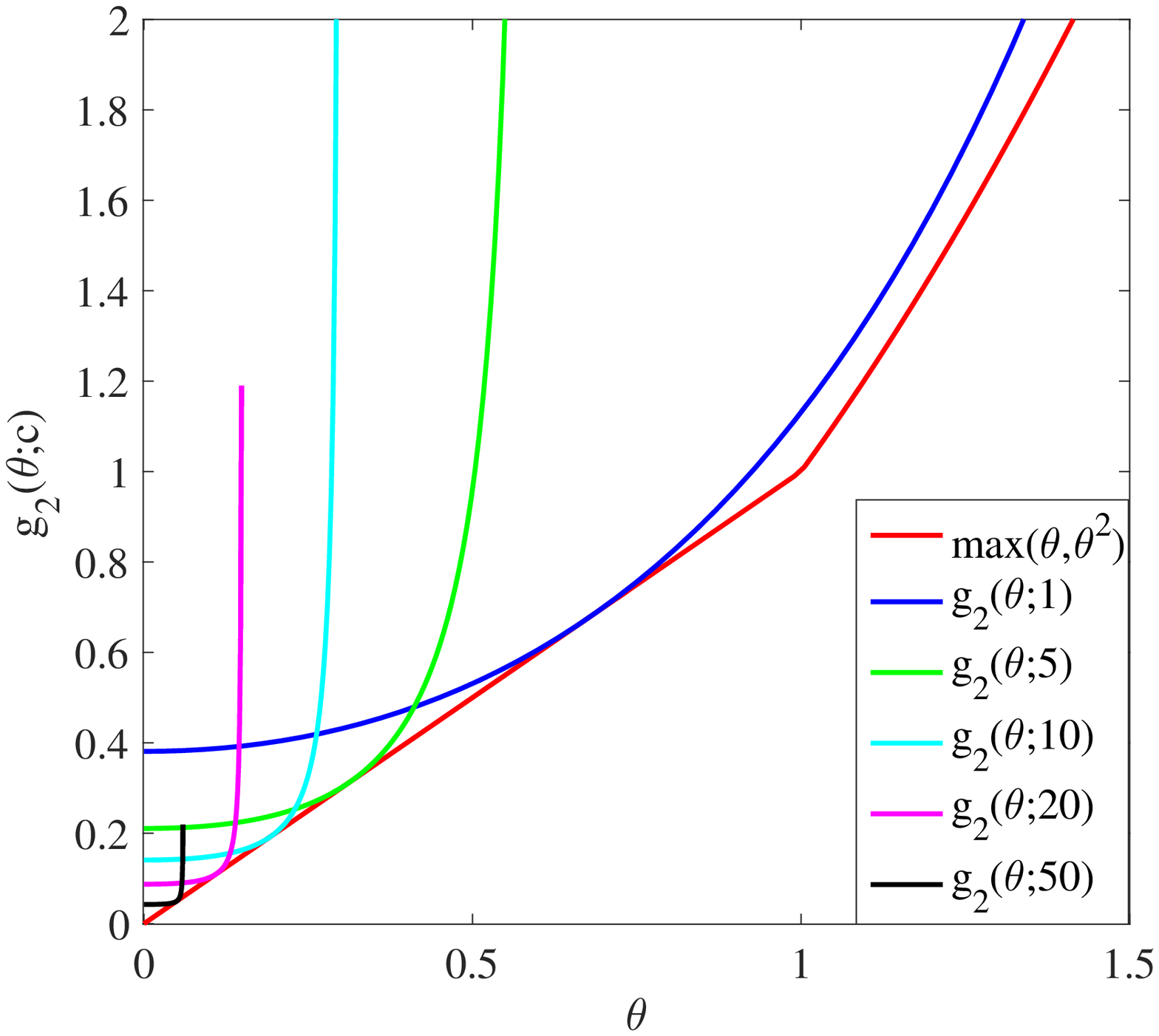} }
\subfigure[\hbox{The curve of $\alpha_2(c)$.}]{
\includegraphics[height=6cm]{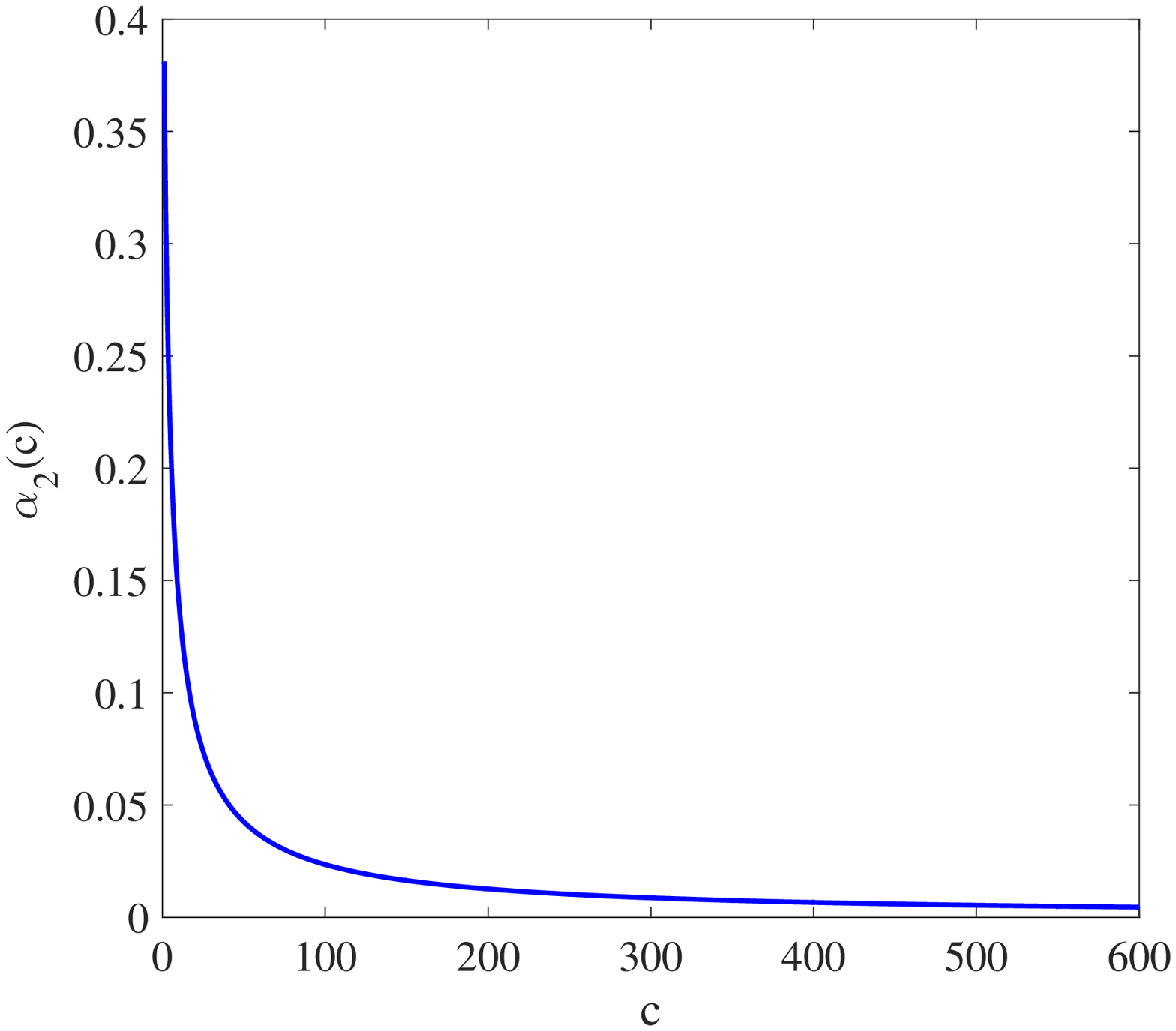}}
\caption{The function curves of $g_2(\theta;c)$ ($c\in\{1,5,10,20,50\}$) and $\alpha_2(c)$.}\label{fig:g3}
\end{figure}

By substituting $g_2(\theta;c)$ and $\phi_{\widetilde{\Omega}}$ into \eqref{eq:exp.bound00}, we then obtain the following expectation bound:
\begin{proposition}\label{prop:exp.bound2}
For any $c>0$, it holds that
\begin{equation}\label{eq:exp.bound3}
\mathbb{E}\left\{ \mu\left(\sum_{k=1}^K{\bf X}_k\right) \right\}\leq \phi_{\widetilde{\Omega}}  \left(\sqrt{2  \alpha_2(c) } +  \frac{c \alpha_2(c)}{3}  \right)  .
\end{equation}
\end{proposition}

Compared with the expectation bounds given in Tropp's works ({\it cf.} Remark 5.5 of \cite{tropp2012user} and Theorem 6.6.1 of \cite{tropp2015introduction}), this result is independent of matrix dimension, and is applicable to various kinds of eigenproblems for sums of random matrices. Subsequently, we will show the applications of this expectation bound in matrix approximation via random sampling.

\subsection{Applications in Matrix Approximation}

For the completeness of presentation, we first introduce the setup of matrix approximation via random sampling and refer to \cite{tropp2015introduction} for further details. Supposed that ${\bf B} \in \mathbb{R}^{m\times n}$ is the objective matrix that can be expressed as the sum of the matrices ${\bf B}_1,\cdots,{\bf B}_L \in \mathbb{R}^{m\times n}$:
\begin{equation*}
{\bf B} = \sum_{l=1}^L {\bf B}_l.
\end{equation*}
We introduce the non-negative quantities $p_1,p_2,\cdots,p_L$ with $\sum_{l=1}^L p_l =1$ to qualify the importance of each summand matrix ${\bf B}_l$, $1\leq l\leq L$. Alternatively, the quantity $p_l$ can also be deemed as the probability with which the corresponding matrix ${\bf B}_l$ is randomly selected in random sampling. The unbiased estimate of the objective matrix ${\bf B}$ can be constructed in the following way: 
\begin{equation*}
{\bf R} = p_l^{-1} {\bf B}_l \quad \mbox{with probability $p_l$},
\end{equation*}
and it is true that $\mathbb{E}\, {\bf R} = \sum_{l=1}^{L} p_l\cdot p_l^{-1} {\bf B}_l = {\bf B}$. 
Although such a random matrix ${\bf R}$ inherits the specific structure of ${\bf B}_l$, it provides a poor approximation of ${\bf B}$ with only a single copy. Thus the average of $K$ independent copies of ${\bf R}$ is adopted to improve the approximation performance, that is,
\begin{equation*}
\widehat{{\bf R}}_K = \frac{1}{K}\sum_{k=1}^K {\bf R}_k.
\end{equation*}
We can select a specific kind of matrix function $\mu(\widehat{{\bf R}}_K - {\bf B})$, such as any matrix norms, as a measurement to examine the approximation performance. The expectation bound \eqref{eq:exp.bound3} leads to the following result on the performance of the matrix approximation via random sampling. 

\begin{theorem}\label{thm:ran.approx}
Assume that the vector space $\mathbb{M}$ and the function $\mu:\mathbb{M}\rightarrow \mathbb{R}$ satisfy Conditions (C1)-(C3). Given a fixed matrix ${\bf B}\in\mathbb{M}$, let the random matrix ${\bf R}\in\mathbb{M}$ be an unbiased estimate of ${\bf B}$. Let ${\bf R}_1,\cdots,{\bf R}_K\in\mathbb{M}$ be the independent copies of ${\bf R}$. Denote $\widehat{{\bf R}}_K := \frac{1}{K}\sum_{k=1}^K {\bf R}_k$ and $u :=\max\limits_{1\leq k\leq K}   \mu( {\bf R}_k - {\bf B})$. If there exists $\epsilon>0$ such that $u \leq  \sqrt{ 1+2\epsilon \mu( {\bf B} ) }-1$, then it holds that for any $c>0$,
\begin{align}\label{eq:ran.approx2}
 \frac{\mathbb{E} \mu( \widehat{{\bf R}}_K - {\bf B})}{\mu({\bf B})} \leq \epsilon \cdot \left(\sqrt{2  \alpha_2(c) } +  \frac{c \alpha_2(c)}{3}  \right).
\end{align}
%
%
\end{theorem}
%
This theorem shows a dimension-free result on the performance of matrix approximation via random sampling.
Especially, when $c$ goes to the {\it infinity}, the term $\big(\sqrt{2  \alpha_2(c) } +  \frac{c \alpha_2(c)}{3}  \big)$ will converge to {\it one}, which means that $ \frac{\mathbb{E} \mu( \widehat{{\bf R}}_K - {\bf B})}{\mu({\bf B})} \leq \epsilon$ in this limiting case. This result highlights the importance of the approximation error $\mu({\bf R}_k - {\bf B})$ caused by each copy ${\bf R}_k$, which suggests that to achieve an accuracy estimate of ${\bf B}$, it should be essential to keep the individual approximation error $\mu({\bf R}_k - {\bf B})$ at a reasonable level. 

\begin{remark}\label{rem:compare}
In Section 6.2 of \cite{tropp2015introduction}, Tropp gave the dimension-dependent result on the approximation error $\mathbb{E}  \| \widehat{{\bf R}}_K - {\bf B}\|$ equipped with the spectral norm $\|\cdot\|$\footnote{This result can be reformulated as $\frac{\mathbb{E}  \| \widehat{{\bf R}}_K - {\bf B}\|}{ \|{\bf B}  \|} \leq O(\epsilon) $ in the applications of matrix random sparsification, randomized matrix multiplication and random feature ({\it cf.} Sections 6.3-6.5 of \cite{tropp2015introduction}).}: for any $\epsilon>0$, it holds that $\mathbb{E}  \| \widehat{{\bf R}}_K - {\bf B}\|\leq 2\epsilon$ if
\begin{equation}\label{eq:tropp}
K \geq \frac{2m_2({\bf R}) \log (m+n)}{\epsilon^2} + \frac{2L \log (m+n)}{3\epsilon},
\end{equation}
where $\mathbf{R}\in\mathbb{C}^{m\times n}$, $\|{\bf R}\|\leq L$ and $m_2({\bf R}) := \max\{ \|\mathbb{E}({\bf R}{\bf R}^*) \|, \|\mathbb{E}({\bf R}^*{\bf R}) \|  \}$. This result suggests that as long as the copy number is large enough, the approximation error can be controlled to be the satisfactory level.
The main differences between the results \eqref{eq:ran.approx2} and \eqref{eq:tropp} lie in the following aspects:
\begin{enumerate}[(1)]
\item Since Tropp's result \eqref{eq:tropp} is dimension dependent, the number $K$ could be very large in the high-dimensional scenario. In contrast, our result is independent of matrix dimension and thus is suitable to the high-dimensional or infinite-dimensional scenario.

\item The bounded condition $\|{\bf R}\|\leq L$ in Tropp' result imposes a requirement into the behavior of the random matrix ${\bf R}$. In contrast, there is no restriction on ${\bf R}$ in our result.

\item Tropp's result is based on the spectral norm, and in contrast, the $\mu(\cdot)$ in our result can be set as variant kinds of matrix functions.

\item Tropp's result shows the asymptotical behavior of the approximation error w.r.t. the copy number $K$. In contrast, our result illustrates a deterministic description to the relationship between the entire and the individual approximation errors.

\end{enumerate}
\end{remark}
To sum up, the two results are complementary with each other. According to Tropp's result, given a quantities of copy matrices ${\bf R}_1,{\bf R}_2,\cdots,{\bf R}_K,\cdots$, the average matrix of any part of them will outperform the individual one and then we treat this average one as a new copy of ${\bf R}$. In this manner, we can generate the series of copy matrices each of which can reach a satisfactory approximate accuracy.


\begin{remark}
Interestingly, there is a direct way to obtain the result $\frac{\mathbb{E} \mu( \widehat{{\bf R}}_K - {\bf B})}{\mu({\bf B})} \leq \epsilon$ instead of the aforementioned limiting case. It begins with the function $g_3(\theta,K) := \theta^{K+1} +\alpha_3(K)$ with  
\begin{equation*}
\alpha_3(K):= \Big(\frac{K}{K+1}\Big)\cdot\Big(\frac{1}{K+1}\Big)^{\frac{1}{K}}.
\end{equation*}
We find that $g_3(\theta,K) \geq  \max\{\theta,\theta^K\}$ for any $\theta \geq 0$ and the curve of $g_3(\theta,K)$ is tangent to that of $\theta$ at the point $\big(\big(\frac{1}{K+1}\big)^{\frac{1}{K}},\big(\frac{1}{K+1}\big)^{\frac{1}{K}}\big)$ ({\it cf.} Fig. \ref{fig:g2}). Substituting $g_3(\theta,K)$ into \eqref{eq:exp.bound00} leads to the expectation bound $\mathbb{E}\big\{ \mu\big(\sum_{k=1}^K{\bf X}_k\big) \big\}\leq  \phi_\Omega$. Similar to Theorem \ref{thm:ran.approx}, if there exists $\epsilon>0$ such that $u \leq   \sqrt[\tau]{ 1+\epsilon \tau \mu( {\bf B} ) }-1$, then it holds that $ \frac{\mathbb{E} \mu( \widehat{{\bf R}}_K - {\bf B})}{\mu({\bf B})} \leq \epsilon.$

\begin{figure}[htbp]
\centering
\subfigure[\hbox{The curves of $g_3(\theta,K)$ and $\max\{ \theta ,\theta^K\}$ ($K=2$).}]{
\includegraphics[height=6cm]{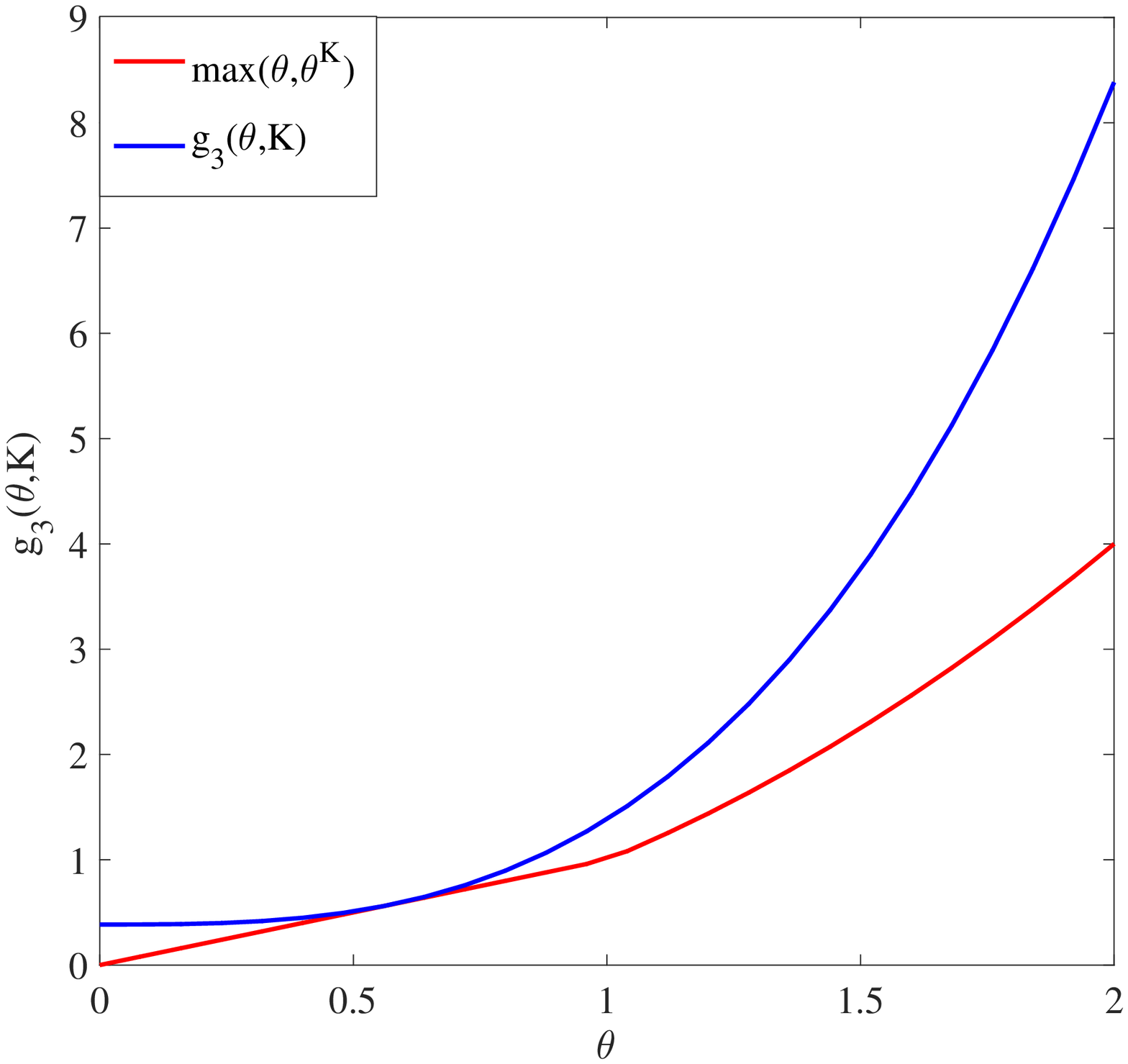} }
\subfigure[\hbox{The curve of $\alpha_3(K)$.}]{
\includegraphics[height=6cm]{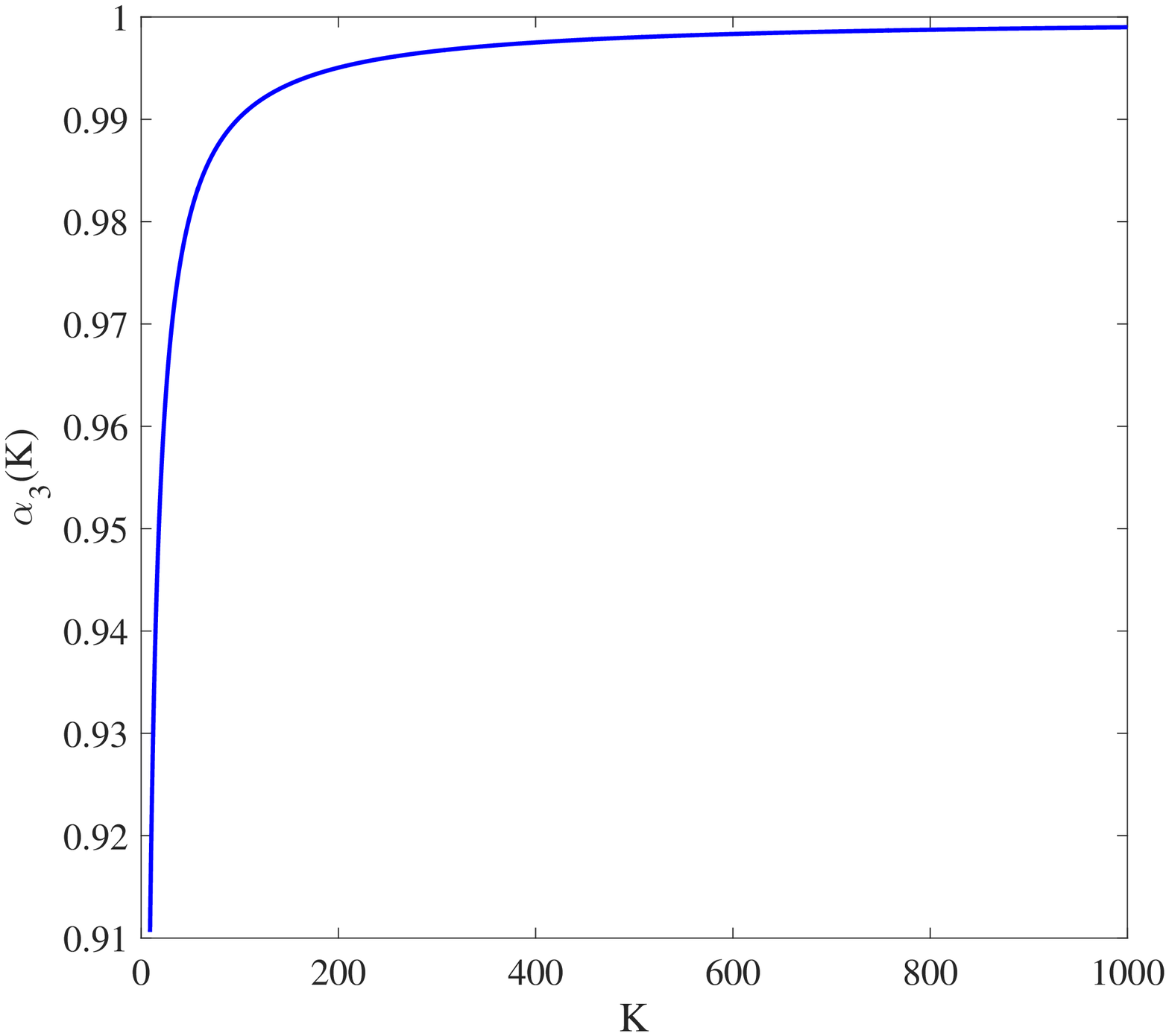}}
\caption{The function curves of $g_3(\theta,K)$ ($K=2$) and $\alpha_3(K)$.}\label{fig:g2}
\end{figure}

\end{remark}


%
%
%


{\section{Applications in Matrix Expander Graphs}\label{sec:sampling}


In this section, we will consider the applications of the proposed framework in quantum information. In particular, we first develop dimension-free tail inequalities for the matrix martingale-difference sequence (MDS). Based on the resulting tail inequality, we then provide a dimension-free analysis to the expander-walk sampling and the cover of quantum hypergraphs.

\subsection{Dimension-free Tail Inequalities for Matrix Martingale Difference Sequence (MDS)}

Given a probability space $(\Omega, \mathcal{F},\mathbb{P})$, denote $\{\mathcal{F}_k\}_{k=0}^{\infty}$ to be a filtration contained in the sigma algebra $\mathcal{F}$, that is, $\mathcal{F}_0 \subset \mathcal{F}_1\subset \mathcal{F}_2\subset \cdots \subset \mathcal{F}_\infty  \subset \mathcal{F}$. By equipping with such a filtration, we define the conditional expectation $\mathbb{E}_k [\cdot] = \mathbb{E}_k [\cdot | \mathcal{F}_k]$. A random-matrix sequence $\{{\bf X}_k\}$is said to be adapted to the filtration if each ${\bf X}_k$ is measurable with respect to $\mathcal{F}_k$. An adapted random-matrix sequence $\{{\bf X}_k\}$ is said to be a matrix martingale if $\mathbb{E}_{k-1}{\bf X}_k = {\bf X}_{k-1}$ and $\mathbb{E} \|{\bf X}_k\|<\infty$ for $k=1,2,3,\cdots$. Given a matrix martingale $\{{\bf X}_k\}$, the matrix martingale difference sequence (MDS) is defined as ${\bf Z}_k := {\bf X}_k - {\bf X}_{k-1}$ for $k=1,2,3,\cdots$. We note that the matrix MDS is conditionally zero mean, that is, $\mathbb{E}_{k-1} {\bf Z}_k = {\bf 0}$.

It is not difficult to verify that the subadditivity of matrix cumulant-generating function \cite[Lemma 3.4]{tropp2012user} still holds for a martingale difference sequence $\{{\bf Z}_1,\cdots,{\bf Z}_K\}$. Then, the result given in Proposition \ref{prop:diagonal.sum} can be extended to the setting of the matrix MDS:
\begin{proposition}\label{prop:diagonal.mds}
Let $\{{\bf Z}_1,\cdots,{\bf Z}_K\}\subset \mathbb{M}$ be a matrix MDS. Then, it holds that for any $\theta>0$,
\begin{align}\label{eq:diagonal.sum}
\mathbb{E}{\rm e}^{\mu\left(\sum_{k=1}^K\theta {\bf Z}_k\right)}\leq {\rm e}^{-1}\cdot {\rm tr}\,\exp\left({\bf D}_0+\sum_{k=1}^K \log\mathbb{E}\,{\rm e}^{{\bf D}_\mu[\theta;{\bf Z}_k]}\right).
\end{align}
\end{proposition}
Similar to the way of developing tail inequalities \eqref{eq:tail2} and \eqref{eq:tail5} for independent matrix sequence, we can derive the dimension-free tail inequalities for the matrix MDS. 
\begin{theorem}\label{thm:tail.mds}
Given a matrix MDS $\{{\bf Z}_1,\cdots,{\bf Z}_K\}\subset \mathbb{M}$, let ${\bf B}_1,\cdots,{\bf B}_K\in\mathbb{M}$ be fixed matrices such that
\begin{equation}\label{eq:cond.b}
\mathbb{E} {\rm e}^{{\bf D}_\mu[\theta;{\bf Z}_k]} \preceq {\rm e}^{{\bf D}_\mu[\theta;{\bf B}_k]},\quad k=1,2,3,\cdots.
\end{equation}
Then, there holds that
\begin{enumerate}
\item for any $\theta>0$,
\begin{align}\label{eq:tail.mds1}
\mathbb{P}\left\{\mu\left(\sum_{k=1}^K{\bf X}_k \right)\geq t \right\}  \leq   {\rm e}^{\varphi_\Omega(1+\alpha_1(\tau))}\cdot\exp\left(- \varphi_\Omega \cdot \Gamma\left(\frac{t}{2\tau\cdot \varphi_\Omega}\right)\right)
\end{align}
where $\varphi_\Omega:=\sum\limits_{i=1}^{I}\big( \big[\mu\big({\bf U}_i\big)+1\big]^{|\Omega_i|}-1\big)$ with ${\bf U}_i = \mathop{\arg\max}\limits_{k\in\Omega_i} \{\mu({\bf B}_k)  \}$.
\item for any $\theta>0$,
\begin{align}\label{eq:tail.mds2}
\mathbb{P}\left\{\mu\left(\sum_{k=1}^K{\bf X}_k \right)\geq t \right\}  \leq  {\rm e}^{\frac{\varphi_{\widetilde{\Omega}}}{4}}\cdot\exp\left\{ -\frac{t^2}{4\phi_{\widetilde{\Omega}}}\right\},
\end{align}
where $\varphi_{\widetilde{\Omega}}:=\sum\limits_{i=1}^{\widetilde{I}}\big( \big[\mu\big(\widetilde{{\bf U}}_i\big)+1\big]^{|\widetilde{\Omega}_i|}-1\big)$ with $\widetilde{{\bf U}}_i = \mathop{\arg\max}\limits_{k\in\widetilde{\Omega}_i} \{\mu({\bf B}_k)  \}$.

\end{enumerate}
\end{theorem}

Subsequently, we will use these tail inequalities to explore the properties of the expander-walk sampling and the cover of quantum hypergraph.

\subsection{Expander-walk Sampling}

The expander-walk sampling refers to a simpler that samples vertices in an expander graph by doing a random walk. It has been proven that such a sampler can generate the samples whose average is not $t$-close to the true mean with exponentially decreasing probability and fewer random bits\cite{gillman1998chernoff}. This finding implies that the sampling results can almost be treated as the independent samples, and thus the expander-walk sampling plays an essential role in quantum information. Although the effectiveness of this method for matrix sampling has been explored in some works \cite{wigderson2005randomness,wigderson2008derandomizing,kyng2018matrix,garg2018matrix}, their results all have the matrix-dimension as the product factor, and thus could not be suitable to the high-dimensional scenario. To overcome this limitation, we will provide the dimension-free analysis of the sampling method under the proposed dimension-free framework.

Given a connected undirected $d$-regular graph $G = (V, E)$ on $n$ vertices, its normalized adjacency matrix ${\bf A}$ is defined as ${\bf A} = [A_{ij}]_{n\times n}$ with $A_{ij} = e_{ij} /d$, where $e_{ij}$ is the number of edges between the $i$-th and the $j$-th vertices. We note that ${\bf A}$ is a real symmetric (certainly Hermitian) matrix and the set of ${\bf A}$'s eigenvalues, called as the spectrum of $G$, is of the form $1 = \lambda_1 > \lambda_2 \geq \lambda_3\geq \cdots \geq \lambda_n$. The unit eigenvector of the eigenvalue $1$ is $(1/\sqrt{n},\cdots,1/\sqrt{n})^T$, and the value of $1-\lambda_2$ is called as the spectral gap of ${\bf A}$. The graph $G = (V, E)$ is said to be an expander graph with spectral gap $\epsilon > 0$ if there holds that $1-\lambda_2>\epsilon$.


Define $y_k$ ($0 \leq k \leq K$) to be the $k$-th vertex visited in a random walk on $G$ and let $\{y_1, \cdots , y_K\}$ be the sequence of vertices encountered on a random walk. A random walk is said to be stationary if it starts from $y_1$ which is chosen uniformly at random. Let 
$f:V \rightarrow \mathbb{H}^{d \times d}$ be a matrix-valued function such that the Frobenius norm $\|f(y)\|_F\leq 1$ for all $y\in V$ and $\sum_{y\in V} f(y) = 0$. Let $\mathbb{E}[f(y)]$ be the mean value of $f(y)$ uniformly over all vertices. Under the assumption that $\mathbb{E}[f(y)] = 0$, we would like to analyze the behavior of the tail probability $\mathbb{P} \big\{ \big\|\frac{1}{K} \sum_{k=1}^K  f(y_k)\big\| >t  \big\}$ ($t>0$). Since the elements of the sequence $\{f(y_1),\cdots,f(y_K)\}$ are not independent of each other yet, the proposed framework cannot be directly used to solve this issue. Instead, the martingale method, proposed by Grag {\it et al.} \cite[Theorem 1.6]{garg2018matrix},  converts the sum of the matrix-valued functions w.r.t. a stationary random walk on an expander graph into the sum of a martingale difference sequence:

\begin{lemma}\label{lem:martingale}
Assume that $\{y_1,\cdots,y_K\}$ is a stationary random walk on the expander graph $G = (V, E)$ with spectral gap $\epsilon > 0$. Then, for any $t>0$, there exists a martingale difference sequence $\{{\bf Z}_1,\cdots,{\bf Z}_K\}$ w.r.t. the filtration generated by initial segments of $\{y_1,\cdots,y_K\}$ such that 
\begin{equation*}
\mathbb{P} \left\{ \left\|\frac{1}{K} \sum_{k=1}^K  f(y_k)\right\| >t  \right\} \leq
\mathbb{P} \left\{ \left\|\frac{1}{K} \sum_{k=1}^K  {\bf Z}_k\right\| >\frac{t}{2}  \right\}. 
\end{equation*}
where ${\bf Z}_k$ is a martingale with bound $\|{\bf Z}_k\| \leq  \frac{\log(n / t)}{\epsilon}$.

\end{lemma}

By combining Theorem \ref{thm:tail.mds}, we then arrive at the Bennett-type and the Azuma-Hoeffding type results, and the latter has the similar form to that of the existing Chernoff bounds for the expander-walk sampling \cite{wigderson2005randomness,wigderson2008derandomizing,kyng2018matrix,garg2018matrix}:

\begin{theorem}\label{thm:expander}
Let $G = (V, E)$ be a expander graph $G = (V, E)$ with spectral gap $\epsilon > 0$. For any $t>0$, $K \geq 1$ and $d\geq 1$, then there exists a ${\rm poly}(r)$-time computable sampler $\sigma : \{0,1\}^r \rightarrow V^K$ with $r = \log(n) + O(K)$ satisfying that 
\begin{enumerate}[(a)]
\item for any $t>0$, 
\begin{align}\label{eq:expander1}
\mathbb{P}_{\omega  \stackrel{R}{\leftarrow}   \{0,1\}^r} \left\{ \left\|\frac{1}{K} \sum_{k=1}^K  f(\sigma(\omega)_k)\right\| >t  \right\} \leq& {\rm e}^{\varphi'_\Omega(1+\alpha_1(\tau))}\cdot\exp\left(- \varphi'_\Omega \cdot \Gamma\left(\frac{Kt}{2\tau\cdot \varphi_\Omega}\right)\right),
\end{align}
where $\omega  \stackrel{R}{\leftarrow}   \{0,1\}^r$ stands for sampling $\omega$ from $\{0,1\}^r$ uniformly, and $\varphi'_\Omega:=\sum\limits_{i=1}^{I}\big( \big[\frac{\log(n / t)}{\epsilon}+1\big]^{|\Omega_i|}-1\big)$;

\item for any $t>0$,
\begin{align}\label{eq:expander2}
\mathbb{P}_{\omega  \stackrel{R}{\leftarrow}   \{0,1\}^r}\left\{\left\|\frac{1}{K} \sum_{k=1}^K  f(\sigma(\omega)_k)\right\| >t \right\}\leq  {\rm e}^{\frac{\varphi'_{\widetilde{\Omega}}}{4}}\cdot\exp\left\{ -\frac{K^2t^2}{16\varphi'_{\widetilde{\Omega}}}\right\},
\end{align}
where $\varphi'_{\widetilde{\Omega}} :=   \sum\limits_{i=1}^{\widetilde{I} }\big( \big[\frac{\log(n / t)}{\epsilon}+1\big]^{|\widetilde{\Omega}_i|}-1\big)  $.

\end{enumerate}

\end{theorem}

\begin{IEEEproof}
As addressed in \cite[Section 5]{goldreich2011sample}, there must exist a sampler via the random walk on the expander graph satisfying the relation $r = \log(n) + O(K)$. Therefore, we only need to prove the inequalities \eqref{eq:expander1} and \eqref{eq:expander2}, which can be directly resulted from the combination of Theorem \ref{thm:tail.mds} and Lemma \ref{lem:martingale}. This completes the proof.
\end{IEEEproof}

Compared with these existing bounds whose product factors are $2d$, our results do not have the matrix dimension as a product factor, and thus can provide a more precise description to the sampler performance when the matrix dimension is high. We note that it follows from the fact $|\widetilde{\Omega}_i| \leq 2$ ($\forall i\in\{1,2,\cdots, | \widetilde{\Omega}|\}$) that  
\begin{equation*}
\varphi'_{\widetilde{\Omega}} \leq  \widetilde{I} \cdot \left[\Big(\frac{\log(n / t)}{\epsilon}\Big)^2+\frac{2\log(n / t)}{\epsilon}\right]= \left\lceil \frac{K}{2} \right\rceil \cdot \left[\Big(\frac{\log(n / t)}{\epsilon}\Big)^2+\frac{2\log(n / t)}{\epsilon}\right].
\end{equation*}
We then obtain a sufficient condition to guarantee the relation ${\rm e}^{\frac{\varphi'_{\widetilde{\Omega}}}{4}}\leq 2d$: 
\begin{equation}\label{eq:choose.k}
K \leq \frac{8\log 2d}{\Big(\frac{\log(n / t)}{\epsilon}\Big)^2+\frac{2\log(n / t)}{\epsilon}}.
\end{equation}
which suggests that the step number $K$ of the random walk should be less than $O(\log 2d)$.

\section{Cover of Quantum Hypergraphs}

We first introduce some necessary preliminaries on quantum hypergraph and refer to \cite[Section 4.3]{wigderson2005randomness} for their details. 

A hypergraph is a pair $(V,E)$ where $E$ is a collection of subsets of $V$. Set $|V|=d$ and an edge $e\in E$ can be treated as a $d \times d$ diagonal matrix with $1$ or $0$ at each diagonal entry to signify whether that vertex is in the edge, where the $i$-th entry is $1$ if the $i$-th vertex is in the edge and $0$ otherwise. Denote the matrix corresponding to the edge $e$ as ${\bf M}_e$. The quantum hypergraph $(\mathcal{V},\mathcal{E})$ is a generalization of the hypergraph generated in the following way: 
\begin{enumerate}[(a)]
\item Let the vertex set $\mathcal{V}$ be a $d$-dimensional complex Hilbert space, and each vertex is represented as a linear combination of an orthonormal basis of $\mathcal{V}$;

\item Given an edge $e\in \mathcal{E}$ containing some vertices in $\mathcal{V}$, the corresponding matrix ${\bf M}_e$ is signified as a projection ${\bf M}_e\in \mathcal{E}$ onto the space spanned by these vertices.

\item For any edge $e\in \mathcal{E}$, the matrix ${\bf M}_e$ is not only limited to the projection, but also extended to be any Hermitian matrix satisfying ${\bf 0} \preceq {\bf M}_e \preceq {\bf I}$.

\end{enumerate}
Therefore, the quantum hypergraph can be formally defined as follows:
\begin{definition}[Quantum Hypergraph]
A hypergraph $G=(\mathcal{V},\mathcal{E})$ is said to be a quantum hypergraph if $\mathcal{V}$ is a $d$-dimensional Hilbert space and $\mathcal{E}$ is a finite set such that each $e \in \mathcal{E}$ is identified with a Hermitian matrix ${\bf M}_e$ with ${\bf 0} \preceq {\bf M}_e \preceq {\bf I}$.
\end{definition}
A finite set $\mathcal{C}\subset \mathcal{E}$ is said to be a cover of a quantum hypergraph $G=(\mathcal{V},\mathcal{E})$ if $\sum_{e\in \mathcal{C}} {\bf M}_e \succeq {\bf I}$. The size of the smallest cover is called the cover number and denoted as ${\rm cov}(G)$. Furthermore, a fractional cover is a set of non-negative weights $w(e)$ ($e\in\mathcal{E}$) such that $\sum_{e\in \mathcal{E}} w(e) {\bf M}_e \succeq {\bf I}$ and the fractional cover number is defined as
\begin{equation*}
{\rm cov}_f(G):= \min_w \left\{ \sum_{e\in \mathcal{E}} w(e) \Big|  \sum_{e\in \mathcal{E}} w(e) {\bf M}_e \succeq {\bf I}    \right\}.
\end{equation*}

One main concern in quantum information is to verify whether the cover of a quantum hypergraph can be found in the polynomial time. This issue has been discussed in the previous works \cite{ahlswede2002strong},\cite[Theorem 4.5]{wigderson2005randomness}, where they mainly concern with the relationship between the cover size and the vertex number (matrix dimension), while the fractional cover number is still treated as a constant. Instead, based on the dimension-free result \eqref{eq:tail5}, we can achieve a new analysis on this issue:

\begin{theorem}\label{thm:graph}
Let $G=(\mathcal{V},\mathcal{E})$ be a quantum hypergraph with the fractional cover number ${\rm cov}_f(G)$ and $|\mathcal{V}|=d$. Then, if ${\rm cov}_f(G) \leq  \frac{K}{6   \left\lceil \frac{K}{2} \right\rceil}$, one can find a $K$-size cover of $G$ in time $d^{K}$.
\end{theorem}
This result shows an upper bound of ${\rm cov}_f(G)$ to guarantee that the cover of $G$ can be found in a polynomial time and illustrates the effect to finding the cover when ${\rm cov}_f(G)$ is super-constant. Our result can be deemed as a complement of the relevant existing works.
We note that this theorem is built on the independent sampling and it could be an interesting problem whether there exists a larger upper bound of ${\rm cov}_f(G)$ when other sampling methods are adopted.

\subsection{Proof of Theorem \ref{thm:graph}}

\begin{IEEEproof} As addressed in the proof of \cite[Theorem 4.5]{wigderson2005randomness}, finding the cover of a quantum hypergraph $G=(\mathcal{V},\mathcal{E})$ can be reduced to a semidefinite program (SDP) problem, and the solving this SDP can provide the fractional cover number in an arbitrary accuracy and a probability distribution of the edges: $p(e) = \frac{w(e)}{{\rm cov}_f(G)}$. Given $p(e)$ and ${\rm cov}_f(G)$, it follows from the definition of ${\rm cov}_f(G)$ that $\mathbb{E}_p[{\bf M}_e] \succeq \frac{1}{{\rm cov}_f(G)}{\bf I}$ and then denote ${\bf M} = \mathbb{E}_p [{\bf M}_e]$. 

Given an i.i.d. sample set $S \subseteq \mathcal{E}$ with $K=|S|$ w.r.t. the distribution $p(e)$, set $K \geq 2{\rm cov}_f(G) $ and it follows from \eqref{eq:tail5} that 
\begin{align}\label{eq:graph.pr1}
\mathbb{P}\left\{  \sum_{e\in S} {\bf M}_e \succeq {\bf I} \right\} = & \mathbb{P}\left\{  \sum_{e\in S} ({\bf M}_e-{\bf M}) \succeq {\bf I} -K {\bf M}\right\}\nonumber\\
\geq &  \mathbb{P}\left\{ \frac{1}{K} \sum_{e\in S} ({\bf M}_e-{\bf M}) \succeq \left(\frac{1}{K} - \frac{1}{{\rm cov}_f(G)}\right){\bf I}  \right\}\nonumber\\
\geq & \mathbb{P}\left\{ \frac{1}{K} \sum_{e\in S} ({\bf M}_e-{\bf M}) \succeq  - \frac{1}{2{\rm cov}_f(G)}{\bf I}  \right\}\nonumber\\
\geq & \mathbb{P}\left\{\left\| \frac{1}{K} \sum_{e\in S} ({\bf M}_e-{\bf M})\right\| \leq \frac{1}{2{\rm cov}_f(G)}  \right\}\nonumber\\
\geq & 1- {\rm e}^{\frac{\varphi_{\widetilde{\Omega}}}{4}}\cdot\exp\left\{ -\frac{K^2}{16\varphi_{\widetilde{\Omega}}{\rm cov}^2_f(G)}\right\}.
\end{align}
On the other hand, since $\|{\bf M}_e-{\bf M}\| \leq 1 $, we have 
\begin{equation}\label{eq:graph.pr2}
\varphi_{\widetilde{\Omega}} \leq 3 \left\lceil \frac{K}{2} \right\rceil.
\end{equation}
To maintain the non-negativity of the above probability, the combination of \eqref{eq:graph.pr1} and \eqref{eq:graph.pr2} leads to
\begin{equation*}
{\rm cov}_f(G) \leq  \frac{K}{6   \left\lceil \frac{K}{2} \right\rceil}.
\end{equation*}
Enumerating over the $K$ i.i.d. samples gives us a deterministic algorithm to find a cover in time $O(d^K)$. This completes the proof.
\end{IEEEproof}

}



\section{Conclusion}

In this paper, we propose a framework to obtain the dimension-free (DF) tail inequalities of a matrix function $\mu$ for sums of random matrices. We also develop the tail inequalities for matrix random series. Although $\mu$ is required to satisfy Conditions (C1-C3), it still contains some usual matrix functions as special cases including {all matrix norms}, the absolute value of sum of the $j$ largest eigenvalues for Hermitian matrices and the sum of $j$ largest singular values for complex matrices. Therefore, the proposed framework can be used to study the tail behavior of many eigenproblems of random matrices. Since the resulted tail inequalities are independent of the matrix dimension, they are suitable to the scenario of high-dimensional or infinite-dimensional matrices. { Compared with the existing works \cite{tropp2012user,zhang2018matrix}, our results are independent of the matrix dimension but also suitable to arbitrary kinds of probability distributions with bounded first-order moment.}

Moreover, we discuss the applications of the resulted dimension-free tail inequalities in the following aspects:

\begin{itemize}
\item In compressed sensing, we achieve a  proof of the restricted isometric property (RIP) for the measurement matrix that can be expressed as the sum of random matrices without any assumption imposed on the distributions of matrix entries. Compared to the previous work \cite{baraniuk2008simple}, instead of the concentration assumption imposed on the measurement matrices, we use the resulted tail inequalities to achieve the proof based on a mild condition \eqref{eq:smallest.sigma} that can be easily satisfied (see Remark \ref{rem:rip.condition}).

\item In probability theory, we bound the supremum of a stochastic process from below. Compared with the existing work \cite{hsu2011dimension}, this upper bound is independent of matrix dimension and is applicable to the various kinds of stochastic processes with unified first-order moments. 

\item In machine learning, we analyze the performance of matrix approximation via random sampling. Our analysis shows that to achieve good approximation, each copy has to approximate the objective matrix well. In contrast, the existing work \cite{tropp2015introduction} highlights the relationship among the matrix dimension, the copy number and the approximation error. 

\item In optimization, the resulted tail inequalities for matrix random series can extend Nemirovski's conjecture \cite{nemirovski2007sums}, which plays an essential part in solving chance constrained optimization problems and quadratic optimization problems with orthogonality constraints, to a more general setting, where the weights can be arbitrary random variables with bounded first-order moments instead of the original condition, that is, either distribution with {\it zero} mean and $[-1,1]$ support or Gaussian distribution with {\it unit} variance ({\it cf.} Remark \ref{rem:series}).

\item {In theoretical computer science, the expander-walk sampling for matrix-valued data plays an essential part, and the effectiveness of this sampling method has become a concerned topic in these years. With help of the random matrix techniques ({\it e.g.} matrix Chernoff bounds), this issue has been studied in many works \cite{wigderson2005randomness,wigderson2008derandomizing,kyng2018matrix,garg2018matrix}. However, their results all have the matrix dimension as a product factor and could become loose when the matrix dimension is high. To overcome this limitation, we provide a dimension-free analysis on the effectiveness of this sampling method.}

\item In quantum information, we analyses the fractional cover number of quantum hypergraphs.  

\end{itemize}

Under the proposed framework, we first obtain the DF tail inequality \eqref{eq:tail1} with the term $\phi$. Since the order of $\phi$ is $O((\mu({\bf{U}}+1))^K)$, this inequality has a rather slow rate of convergence to {\it zero} in the case of large $K$. To overcome this issue, we present  the tail inequality \eqref{eq:tail2} that is equipped with the term $\phi_\Omega$. Since the order of $\phi_\Omega$ is much lower than that of $\phi$, the inequality \eqref{eq:tail2} can converge to {\it zero} at a reasonable rate in spite of large $K$. The experimental results support the validity of the proposed framework and show that the inequality \eqref{eq:tail2} provides a better description to the tail behavior of the probability $\mathbb{P}\left\{\mu\left(\sum_k{\bf X}_k \right)\geq t \right\}$. 
In the future works, we will explore further applications of the resulted tail inequalities.





\appendix

\section{Proof of Main Results}

In the appendix, we give the proofs of Proposition \ref{prop:operation}, Proposition \ref{prop:diagonal}, Proposition \ref{prop:diagonal.sum}, Proposition \ref{prop:master1}, {Theorem \ref{thm:tail1}, Proposition \ref{prop:validity}}, Theorem \ref{thm:select}, Theorem \ref{thm:rip}, {Lemma \ref{lem:exp.bound}, Proposition \ref{prop:exp.bound2} and Theorem \ref{thm:ran.approx},} respectively.

\subsection{Proof of Proposition \ref{prop:operation}}

\begin{IEEEproof} (1) It follows from Condition (C3) that 
\begin{equation}\label{eq:operation.pr1}
\mu\left(\theta\cdot\sum_{k=1}^K{\bf B}_{k}\right)  \leq \sum_{k=1}^K\mu(\theta\cdot{\bf B}_k).
\end{equation}
Thus, we have
\begin{align}\label{eq:operation.pr2}
{\bf D}_\mu\left[\theta;\sum_{k=1}^K {\bf B}_k\right]
=& \bm{\Lambda}\left[0, \log\left(\mu\left(\theta\cdot\sum_{k=1}^K {\bf B}_k\right)+1\right),2\log\left(\mu\left(\theta\cdot\sum_{k=1}^K {\bf B}_k\right)+1\right), \cdots\right]\nonumber\\
\leq & \sum_{k=1}^K\bm{\Lambda}\Big[0, \log\big(\mu(\theta\cdot{\bf B}_k)+1\big),2\log\big(\mu(\theta\cdot{\bf B}_k)+1\big), \cdots\Big]\nonumber\\
=&\sum_{k=1}^K {\bf D}_\mu[\theta;{\bf B}_k],
\end{align}
where the inequality is derived from the fact that
\begin{equation*}
\log (x+y+1)\leq \log (x+1)+\log(y+1),\;\;\forall\;x,y>0.
\end{equation*}
This leads to the inequality \eqref{eq:operation1}.

(2) According to the inequality of Arithmetic and geometric means: 
\begin{equation*}
\frac{s_1+s_2+\cdots+s_K}{K} \geq \sqrt[K]{s_1  s_2  \cdots  s_K},\quad \forall\;s_1,\cdots,s_K \geq 0,
\end{equation*}
we have for any $\theta>0$,
\begin{equation}\label{eq:inequality.fk}
\sum_{k=1}^K \log (\theta \cdot s_k+1) \leq K\cdot \log \left(  1+\sum_{k=1}^K \frac{\theta \cdot s_k}{K} \right).
\end{equation}
The combination of Condition (C2), \eqref{eq:def.diag1} and \eqref{eq:inequality.fk} leads to
\begin{align*}
\sum_{k=1}^K{\bf D}_\mu[\theta; {\bf B}_k]\preceq K\cdot {\bf D}_\mu\left[\theta;\sum_{k=1}^K\frac{{\bf A}_k}{K}\right].
\end{align*}
The partial super-additivity \eqref{eq:operation3} can be proved in the same way, so we omit it here. This completes the proof. 
\end{IEEEproof}

\subsection{Proof of Proposition \ref{prop:diagonal}}

\begin{IEEEproof} (1) Given a matrix ${\bf B}\in\mathbb{M}$ and $\theta>0$, denote $s:=\mu({\bf B})$. It follows from Taylor's expansion ${\rm e}^{x}=1+\sum\limits_{k=1}^\infty \frac{x^k}{k!}$ that
\begin{align}\label{eq:diagonal.pr2}
{\rm e}^{\theta s+1}
= {\rm tr}\Big(\bm{\Lambda}\Big[1,\theta s+1,\frac{(\theta s+1)^2}{2!}, \frac{(\theta s+1)^3}{3!},\cdots\Big] \Big).
\end{align}
The combination of \eqref{eq:def.diag}, \eqref{eq:def.diag0}, \eqref{eq:def.diag1} and \eqref{eq:diagonal.pr2} leads to
\begin{equation*}
{\rm e}^{\mu(\theta{\bf B})}={\rm e}^{\theta \mu({\bf B})}={\rm e}^{-1}\cdot {\rm e}^{\theta \mu({\bf B})+1}={\rm e}^{-1}\cdot {\rm tr}\big( {\rm e}^{{\bf D}_\mu[\theta; {\bf B}]+{\bf D}_0}\big) = {\rm e}^{-1}\cdot {\rm tr}\big( {\rm e}^{\widehat{{\bf D}}_\mu[\theta; {\bf B}]}\big).
\end{equation*}

(2) According to \eqref{eq:def.diag1}, we have 
\begin{align*}
K\cdot  {\bf D}_\mu[\theta; {\bf B}] =& K \cdot \bm{\Lambda}\Big[0, \log(\theta \cdot \mu({\bf B})+1),2\log(\theta \cdot \mu({\bf B})+1), 3\log(\theta \cdot \mu({\bf B})+1),\cdots\Big]\nonumber\\
=& \bm{\Lambda}\Big[0, \log\big((\theta \cdot \mu({\bf B})+1)^K\big),2\log\big((\theta \cdot \mu({\bf B})+1)^K\big), 3\log\big((\theta \cdot \mu({\bf B})+1)^K\big),\cdots\Big].
\end{align*}
In the similar way, it also follows from Taylor's expansion of ${\rm e}^{x}$ that 
\begin{align*}
{\rm e}^{(\mu(\theta \cdot {\bf B})+1)^K}={\rm tr}\,{\rm e}^{{\bf D}_0+K\cdot {\bf D}_\mu[\theta; {\bf B}]}.
\end{align*}
This completes the proof. 
\end{IEEEproof}


\subsection{Proof of Proposition \ref{prop:diagonal.sum}}

\begin{IEEEproof} It follows from the sub-additivity \eqref{eq:operation1} that 
\begin{align}\label{eq:diagonalsum.pr3}
{\rm tr}\, \exp\left({\bf D}_\mu\left[\theta;\sum_{k=1}^K {\bf X}_k\right]\right)
\leq&{\rm tr}\,\exp\left(\sum_{k=1}^K {\bf D}_\mu[\theta;{\bf X}_k]
\right).
\end{align}
According to \cite[Lemma 3.4]{tropp2012user}, it holds that
\begin{align}\label{eq:diagonalsum.pr4}
\mathbb{E}\,{\rm tr}\,\exp\left({\bf D}_0+\sum_{k=1}^K {\bf D}_\mu[\theta;{\bf X}_k]\right)
\leq  {\rm tr}\,\exp\left({\bf D}_0+\sum_{k=1}^K \log\mathbb{E}\,{\rm e}^{{\bf D}_\mu[\theta;{\bf X}_k]}\right).
\end{align}
By combining \eqref{eq:diagonalsum.pr3}, \eqref{eq:diagonalsum.pr4} and Proposition \ref{prop:diagonal}, we have
\begin{align}
\mathbb{E}\,\exp\left(\mu\left(\theta\cdot \sum_{k=1}^K{\bf X}_k\right)\right)= & {\rm e}^{-1} \cdot \mathbb{E}\,{\rm tr} \,\exp\left(\widehat{{\bf D}}_\mu\left[\theta;\sum_{k=1}^K{\bf X}_k\right]\right)\nonumber\\
=& {\rm e}^{-1} \cdot \mathbb{E}\,{\rm tr} \,\exp\left({\bf D}_0 +{\bf D}_\mu\left[\theta;\sum_{k=1}^K{\bf X}_k\right]\right)\nonumber\\
\leq & {\rm e}^{-1} \cdot \mathbb{E}\,{\rm tr} \,\exp\left({\bf D}_0+\sum_{k=1}^K{\bf D}_\mu\left[\theta;{\bf X}_k\right]\right)\nonumber\\
%
%
\leq &{\rm e}^{-1} \cdot{\rm tr}\,\exp\left({\bf D}_0+\sum_{k=1}^K \log\mathbb{E}\,{\rm e}^{{\bf D}_\mu[\theta;{\bf X}_k]}\right).\label{eq:diagonalsum.pr5}
\end{align}
This completes the proof. 

\end{IEEEproof}

\subsection{Proof of Proposition \ref{prop:master1}}

\begin{IEEEproof} Since $a^K-1 = (a-1 )(\sum_{k=0}^{K-1}a^{k})$, we have for any $s,\theta>0$,
\begin{align}
(\theta s+1)^K -1
=& \theta s \left(\sum_{k=0}^{K-1}(\theta s+1)^k \right)\nonumber\\
 \leq&\left\{ \begin{array}{ll}
 \theta \cdot \left((s+1)^K-1 \right), &\mbox{if\; $0<\theta\leq 1$;} \\
 \theta \cdot\theta^{K-1}\cdot \left((s+1)^K-1 \right), & \mbox{if\; $\theta > 1$,}
\end{array}\right.\nonumber\\
\leq & g(\theta,K)\cdot \left((s+1)^K-1 \right), \label{eq:g_function}
\end{align}
where $g(\theta,K)$ is any function satisfying $g(\theta,K) \geq \max\{\theta,\theta^K\}$ for $\theta\geq 0$.

Combining Markov's inequality and the super-additivity \eqref{eq:operation2} yields, for any $\theta>0$,
\begin{align}\label{eq:laps.pr2}
\mathbb{P}\left\{ \mu\left(\sum_{k=1}^K{\bf X}_k\right)\geq t \right\}=&\mathbb{P}\left\{  \mu\left(\theta \cdot\sum_{k=1}^K{\bf X}_k\right)\geq \theta t \right\}\nonumber\\
=&\mathbb{P}\left\{\exp\left( \mu\left(\theta\cdot \sum\limits_{k=1}^K{\bf X}_k\right)\right)\geq {\rm e}^{\theta t} \right\}\nonumber\\
\leq& {\rm e}^{-\theta t}\cdot \mathbb{E}\,\exp\left( \mu\left(\theta\cdot \sum\limits_{k=1}^K{\bf X}_k\right)\right)\nonumber\\
\leq & {\rm e}^{-\theta t}\cdot{\rm e}^{-1}\cdot {\rm tr}\,\exp\left({\bf D}_0+\sum_{k=1}^K \log\mathbb{E}\,{\rm e}^{{\bf D}_\mu[\theta;{\bf X}_k]}\right)\nonumber\\
\leq & {\rm e}^{-\theta t}\cdot{\rm e}^{-1}\cdot {\rm tr}\,\exp\left( {\bf D}_0+\sum_{k=1}^K \log\,{\rm e}^{{\bf D}_\mu[\theta;{\bf B}_k] }\right)\nonumber\\
=& {\rm e}^{-\theta t}\cdot{\rm e}^{-1}\cdot {\rm tr}\,\exp\left({\bf D}_0+\sum_{k=1}^K {\bf D}_\mu[\theta;{\bf B}_k]\right)\nonumber\\
\leq & {\rm e}^{-\theta t}\cdot{\rm e}^{-1}\cdot {\rm tr}\,\exp\left( K\cdot {\bf D}_\mu\left[\theta;{\bf U}\right] +{\bf D}_0\right)\nonumber\\
\leq &  \exp \big(-\theta t + g(\theta,K) \cdot \phi \big),
\end{align}
where $\phi:=\left[\mu\left( {\bf U}\right)+1\right]^K-1$ with ${\bf U} = \mathop{\arg\max}\limits_{1\leq k\leq K} \{ \mu ({\bf B}_k)\}$ and $g(\theta,K) \geq \max\{\theta,\theta^K\}$. Equation (\ref{eq:laps.pr2}) follows from (\ref{eq:diagonal.2}) and (\ref{eq:g_function}).
Finally, taking an infimum w.r.t. $\theta$ completes the
proof.
\end{IEEEproof}

{\subsection{Proof of Theorem \ref{thm:tail1}}

\begin{IEEEproof}
By substituting $g_1(\theta,K)$ into the right-hand side of the master inequality \eqref{eq:master1}, we have
\begin{align}\label{eq:tail1.pr1}
\mathbb{P}\left\{\mu\left(\sum_{k=1}^K{\bf X}_k \right)\geq t \right\} \leq \inf_{\theta >0} \left\{{\rm e}^{-\theta t+ \phi ( {\rm e}^{K \theta} -K \theta + \alpha_1(K))}  \right\}.
\end{align}
Denote $h(\theta) : = -\theta t+ \phi ( {\rm e}^{K \theta} -K \theta + \alpha_1(K))$. The solution to the equation $\frac{d h(\theta)}{d \theta} = 0$ is  
\begin{align*}
\theta = \frac{1}{K} \log\left(1 + \frac{t}{\phi K}\right),
\end{align*}
which minimizes $h(\theta)$ over all $\theta>0$ with the minimization
\begin{align*}
\min_{\theta>0} \{h(\theta) \}= -\phi \left[  \left(1 + \frac{t}{\phi K}\right)   \log\left(1 + \frac{t}{\phi K}\right)  -\frac{t}{\phi K} \right] + \phi(  \alpha_1(K) +1).
\end{align*}
Setting $\Gamma(x):= (1+x)\log (1+x)-x$ ($x>0$) leads to the first inequality in \eqref{eq:tail1}. The last two inequalities of \eqref{eq:tail1} are derived from the fact that 
\begin{equation*}
\Gamma(x) \geq \frac{x^2}{2(1+x/3)} \geq \left\{
\begin{array}{ll}
 \frac{3x}{4}, & x\geq 3;      \\
   \frac{x^2}{4}, &  0< x <3. 
\end{array}
\right.
\end{equation*}
This completes the proof.
\end{IEEEproof}
}

\subsection{Proof of Proposition \ref{prop:validity}}

\begin{IEEEproof}
According to \eqref{eq:def.diag1}, for any $\theta>0$, we have
\begin{align*}
\mathbb{E}\,{\rm e}^{{\bf D}_\mu[\theta;{\bf X}]} := \bm{\Lambda}[1, \mathbb{E}(\theta\mu({\bf X})+1), \mathbb{E}(\theta\mu({\bf X})+1)^2, \mathbb{E}(\theta\mu({\bf X})+1)^3,\cdots   ],
\end{align*}
and 
\begin{equation*}
{\rm e}^{{\bf D}_\mu[\theta;{\bf B}]} := \bm{\Lambda}[1, (\theta\mu({\bf B})+1), (\theta\mu({\bf B})+1)^2, (\theta\mu({\bf B})+1)^3,\cdots   ].
\end{equation*}
Thus, we only need to prove that if $\mathbb{E}\mu({\bf X})\leq \mu({\bf B})$, the inequality $\mathbb{E}(\theta\mu({\bf X})+1)^n \leq (\theta\mu({\bf B})+1)^n$ holds for any $n\geq 2$. In fact, since $\mu({\bf X}) \geq 0$ and
\begin{equation*}
a^n - b^n = (a-b)\left( \sum_{i=0}^{n-1} a^{i}b^{n-1-i}\right),\quad n\geq 2,
\end{equation*}
we have
\begin{align*}
&(\theta\mu({\bf B})+1)^n - \mathbb{E}(\theta\mu({\bf X})+1)^n \nonumber\\
=& \mathbb{E} \Big[  (\theta\mu({\bf B})+1)^n - (\theta\mu({\bf X})+1)^n  \Big]\nonumber\\
= & \mathbb{E} \left[\theta\big(\mu({\bf B})- \mu({\bf X})\big)\cdot \left( \sum_{ i=0}^{n-1}  (\theta\mu({\bf B})+1)^i  (\theta\mu({\bf X})+1)^{n-i-1}   \right)   \right]\nonumber\\
\geq & \mathbb{E} \left[\theta\big(\mu({\bf B})- \mu({\bf X})\big)\cdot \left( \sum_{ i=0}^{n-1}  (\theta\mu({\bf B})+1)^i  \right)   \right]\nonumber\\
 = & \theta  \left( \sum_{ i=0}^{n-1}  (\theta\mu({\bf B})+1)^i  \right) \cdot  \big(\mu({\bf B})- \mathbb{E} \mu({\bf X})   \big)
 \geq 0.
\end{align*}
This completes the proof.
\end{IEEEproof}


\subsection{Proof of Theorem \ref{thm:select}}

\begin{IEEEproof}
According to \eqref{eq:bgamma}, by setting 
\begin{equation*}
\gamma' =\gamma\cdot \exp\left(\frac{1}{N}\sum_{n=1}^N \log \big(\mu(\theta {\bf X}^{(n)})+1\big) \right),
\end{equation*}
we have for any $\theta>0$
\begin{align}\label{eq:select.pr1}
 \log (\mu( \theta{\bf B}_\gamma)+1) \geq &\log  \left(\theta \left( \frac{\mu({\bf X}^{(1)})+\cdots+\mu({\bf X}^{(N)})}{N}+\gamma' \right)+1\right)\nonumber\\
\geq & \log  \left(\theta \left( \sqrt[N]{\prod_{n=1}^N\Big(\mu({\bf X}^{(n)})+\frac{1}{\theta}\Big)}
- \frac{1}{\theta}
+\gamma' \right)+1\right)\quad (*)\nonumber\\
= & \log  \left(\theta \left(\frac{1}{\theta}\exp\left(\frac{1}{N}\sum_{n=1}^N \log \big(\mu(\theta {\bf X}^{(n)})+1\big) \right)- \frac{1}{\theta}
+\gamma' \right)+1\right)\nonumber\\
= & \log  \left( \exp\left(\frac{1}{N}\sum_{n=1}^N \log \big(\mu(\theta {\bf X}^{(n)})+1\big) \right)
+\theta\gamma'\right)\nonumber\\
= & \frac{1}{N}\sum_{n=1}^N \log \big(\mu(\theta {\bf X}^{(n)})+1\big) +\log(1+\theta \gamma ),
\end{align}
where the $(*)$ step is resulted from the fact that $\sqrt[N]{x_1x_2\cdots x_N}\leq \frac{x_1+x_2+\cdots+x_N}{N}$ ($\forall\,x_1,\cdots,x_N\geq 0$). By using \eqref{eq:select.pr1}, we then arrive at 
\begin{align*}
&\mathbb{P} \left\{   \mathbb{E}\log \big(\mu(\theta {\bf X})+1\big)> \log \big(\mu(\theta {\bf B}_\gamma)+1\big)     \right\}\nonumber\\
=&\mathbb{P} \left\{   \mathbb{E}\log \big(\mu(\theta {\bf X})+1\big)-  
\frac{1}{N}\sum_{n=1}^N \log \big(\mu(\theta {\bf X}^{(n)})+1\big) \right. \nonumber\\
 &\qquad\qquad\left. > \log \big(\mu(\theta {\bf B}_\gamma)+1\big) -\frac{1}{N}\sum_{n=1}^N \log \big(\mu(\theta {\bf X}^{(n)})+1\big)     \right\}\nonumber\\
\leq &\mathbb{P} \left\{   \mathbb{E}\log \big(\mu(\theta {\bf X})+1\big)-  
\frac{1}{N}\sum_{n=1}^N \log \big(\mu(\theta {\bf X}^{(n)})+1\big)    > \log(1+\theta \gamma )     \right\}\nonumber\\
\leq& \exp\left( \frac{-N (\log(1+\theta \gamma ) )^2}{2 \mathbb{E} \big(\log \big(\mu(\theta {\bf X})+1\big)\big)^2     }   \right),
\end{align*}
where the last inequality follows from  \cite[Theorem 2.7]{chung2006complex}. This completes the proof.
\end{IEEEproof}

\subsection{Proof of Theorem \ref{thm:rip}}

In order to prove Theorem \ref{thm:rip}, we first need a preliminary lemma as follows:

\begin{lemma}\label{lem:rip}
Let ${\bf P}_1,\cdots,{\bf P}_K \in \mathbb{C}^{m\times s}$ be random matrices and ${\bf P} =\sum_{k=1}^K  {\bf P}_k$ satisfying  
\begin{equation}\label{eq:smallest.sigma1}
\sigma_{\min} ({\bf P}) \geq \frac{\sigma_{\min} ({\bf P}_1)+\cdots + \sigma_{\min} ({\bf P}_K) }{K^2}.
\end{equation}
Let $\{{\bf A}_1,\cdots,{\bf A}_K\}$ and $\{{\bf B}_1,\cdots,{\bf B}_K\}$ be two matrix sequences such that 
\begin{align*}
\mathbb{E}\,{\rm e}^{{\bf D}_{\sigma_{\max}}[\theta;{\bf P}_k]} \preceq {\rm e}^{{\bf D}_{\sigma_{\max}}[\theta;{\bf A}_k]}\;\; \mbox{and}\;\; \mathbb{E}\,{\rm e}^{{\bf D}_{\sigma_{\max}}[\theta;{\bf P}^{\dag}_k]} \preceq {\rm e}^{{\bf D}_{\sigma_{\max}}[\theta;{\bf B}_k]}
\end{align*}
hold for any $1\leq k\leq K$. Let $\Omega = \{\Omega_1,\cdots,\Omega_I\}$ be a partition of the index set $\{1,\cdots, K\}$ with $\bigcup_{i=1}^I \Omega_i=\{1,\cdots, K\}$, and $\tau: = \max\limits_{1\leq i\leq I}\{ |\Omega_i|\}$. Denote $\bar{\phi}_0 := \max\{\bar{\phi}_1,\bar{\phi}_2\}$ with
\begin{equation*}
\bar{\phi}_1 := \sum_{i=1}^I\big [(\sigma_{\max}({\bf V}_{i})+1)^{|\Omega_i|}-1\big] \;\; \mbox{and}\;\; \bar{\phi}_2 := \sum_{i=1}^I \big [(\sigma_{\max}({\bf U}_i)+1)^{|\Omega_i|}-1\big],
\end{equation*}
where ${\bf V}_{i} := \mathop{\arg\max}\limits_{k\in\Omega_i}\{  \sigma_{\max}({\bf A}_{k}) \}$ and ${\bf U}_i := \mathop{\arg\max}\limits_{k\in\Omega_i}\{  \sigma_{\max}({\bf B}_{k}) \}$. Then, for any $0<\delta <1$, it holds that
\begin{equation}\label{eq:cs.lem1}
(1-\delta) \| {\bf x}  \|_2 \leq \| {\bf P}{\bf x}    \|_2 \leq (1+\delta) \| {\bf x} \|_2\quad  ({\bf x}\in \mathbb{R}^s)
\end{equation}
with probability at least 
\begin{equation*}
1- 2\exp\left(- \bar{\phi}_0\cdot \Gamma\left(\frac{t}{\tau\cdot\bar{\phi}_0}\right)\right),
\end{equation*}
where $\Gamma(t):= (t+1)  \log(t+1) - t$.\end{lemma}

\begin{IEEEproof} {For any $c>0$, the term ${\rm e}^{- \phi \cdot \Gamma(\frac{c}{\phi})}$ is an increasing function w.r.t. $\phi>0$, because 
\begin{align*}
\frac{{\rm d} }{{\rm d} \phi} {\rm e}^{- \phi \cdot \Gamma(\frac{c}{\phi})} = \left(\frac{c}{\phi}-\ln\left(\frac{c}{\phi}+1\right)\right) \cdot {\rm e}^{- \phi \cdot \Gamma(\frac{c}{\phi})}> 0, \quad \forall \phi>0.
\end{align*}
Thus, it follows from \eqref{eq:tail2} that 
\begin{align*}
&\mathbb{P}\left\{ \|{\bf P} {\bf x}\|_2 > (1+\delta) \|{\bf x}\|_2 ,\;\; \forall \, {\bf x}\in\mathbb{R}^s  \right\}\nonumber\\
 = & \mathbb{P} \left\{ \sigma_{\max}({\bf P}) >   (1+\delta) \right\}\nonumber\\
  \leq &{\rm e}^{(1+\alpha_1(\tau))\bar{\phi}_1}\cdot\exp\left(- \bar{\phi}_1\cdot \Gamma\left(\frac{1+\delta}{\tau\cdot \bar{\phi}_1}\right)\right)\nonumber\\
  \leq &{\rm e}^{(1+\alpha_1(\tau))\bar{\phi}_0}\cdot\exp\left(- \bar{\phi}_0\cdot \Gamma\left(\frac{1+\delta}{\tau\cdot \bar{\phi}_0}\right)\right).
\end{align*}}
On the other hand, since $\sigma_{\min}({\bf P}) = \frac{1}{ \sigma_{\max} ({\bf P}^\dag)}$ and $\frac{x_1+\cdots+x_K}{K} \geq \frac{K}{\frac{1}{x_1}+\cdots+ \frac{1}{x_K}}$ ($x_1,\cdots,x_K >0$), it follows from Condition \eqref{eq:smallest.sigma1} that
\begin{align*}
\sigma_{\max}({\bf P}^\dag) \leq \sigma_{\max}({\bf P}_1^\dag) +\cdots + \sigma_{\max}({\bf P}_K^\dag).
\end{align*}
Since $(1+\delta)< (1-\delta)^{-1}$ holds for any $\delta\in(0,1)$, we also have 
\begin{align*}
&\mathbb{P}\left\{ \|{\bf P} {\bf x}\|_2 < (1-\delta) \|{\bf x}\|_2 ,\;\; \forall \, {\bf x}\in\mathbb{R}^s  \right\}\nonumber\\
 = & \mathbb{P} \left\{ \sigma_{\min}({\bf P}) <   (1-\delta) \right\}\nonumber\\
 = & \mathbb{P} \left\{ \sigma_{\max}({\bf P}^\dag) >   (1-\delta)^{-1} \right\}\nonumber\\
 \leq &  \mathbb{P} \left\{ \sum_{k=1}^K \sigma_{\max}({\bf P}_k^\dag) >   (1-\delta)^{-1} \right\}\nonumber\\
  \leq & {\rm e}^{(1+\alpha_1(\tau))\bar{\phi}_2}\cdot\exp\left(- \bar{\phi}_2\cdot \Gamma\left(\frac{(1-\delta)^{-1}}{\tau\cdot \bar{\phi}_2}\right)\right)\nonumber\\
  \leq & {\rm e}^{(1+\alpha_1(\tau))\bar{\phi}_0}\cdot\exp\left(- \bar{\phi}_0\cdot \Gamma\left(\frac{(1+\delta)}{\tau\cdot \bar{\phi}_0}\right)\right).
\end{align*}
This completes the proof.
\end{IEEEproof}

Now, we come up with the proof of Theorem \ref{thm:rip}:
\begin{IEEEproof}
As shown in Lemma \ref{lem:rip}, for each $\mathcal{I} \subset \{1,\cdots,D\}$ with $|\mathcal{I}| =s$, the $m\times s$ random matrix $[{\bf P}]_\mathcal{I} = \sum_{k=1}^K [{\bf P}_k]_\mathcal{I}$ fails to satisfy the RIP \eqref{eq:cs.lem1} with probability at most
\begin{equation*}
2 {\rm e}^{(1+\alpha_1(\tau))\bar{\phi}_\Omega}\cdot \exp\left(- \bar{\phi}_\Omega \cdot \Gamma\left(\frac{t}{\tau\cdot \bar{\phi}_\Omega}\right)\right).
\end{equation*}
Since there are ${n \choose s}\leq ({\rm e} n/s)^s$ possibilities to select $\mathcal{I}$ from $\{1,\cdots,n\}$, the ${\rm RIP_s}(\delta)$ \eqref{eq:rip1} will fail to hold with probability at most 
\begin{equation}\label{eq:rip.pr2}
2{\rm e}^{(1+\alpha_1(\tau))\bar{\phi}_\Omega}\cdot ({\rm e} n/s)^s\cdot\exp\left(- \bar{\phi}_\Omega\cdot \Gamma\left(\frac{t}{\tau\cdot \bar{\phi}_\Omega}\right)\right).
\end{equation}
Therefore, if the constants $c_1,c_2>0$ satisfy Conditions \eqref{eq:rip.cond1} and \eqref{eq:rip.cond2}, then the expression \eqref{eq:rip.pr2} will be smaller than $2{\rm e}^{(1+\alpha_1(\tau))\bar{\phi}_\Omega}\cdot{\rm e}^{-c_2m}$. This completes the proof. 
\end{IEEEproof}

%
%

\subsection{Proof of Lemma \ref{lem:exp.bound}}   

\begin{IEEEproof}
Continuing from the inequality (\ref{eq:diagonalsum.pr5}), we have 
\begin{align*}
\mathbb{E}\,\exp\left(\mu\left(\theta\cdot \sum_{k=1}^K{\bf X}_k\right)\right)= & {\rm e}^{-1} \cdot \mathbb{E}\,{\rm tr} \,\exp\left(\widehat{{\bf D}}_\mu\left[\theta;\sum_{k=1}^K{\bf X}_k\right]\right)\nonumber\\
\leq &{\rm e}^{-1} \cdot{\rm tr}\,\exp\left({\bf D}_0+\sum_{k=1}^K \log\mathbb{E}\,{\rm e}^{{\bf D}_\mu[\theta;{\bf X}_k]}\right).\label{eq:diagonalsum.pr5}
\end{align*}
Then
\begin{align}
\mathbb{E}\,\exp\left(\mu\left(\theta\cdot \sum_{k=1}^K{\bf X}_k\right)\right)
%
\leq &\ {\rm e}^{-1}\cdot {\rm tr}\,\exp\left( {\bf D}_0+\sum_{k=1}^K \log\,{\rm e}^{{\bf D}_\mu[\theta;{\bf B}_k] }\right)\nonumber\\
=&\ {\rm e}^{-1}\cdot {\rm tr}\,\exp\left({\bf D}_0+\sum_{k=1}^K {\bf D}_\mu[\theta;{\bf B}_k]\right)\nonumber\\
\leq &\ {\rm e}^{-1}\cdot {\rm tr}\,\exp\left( K\cdot {\bf D}_\mu\left[\theta;{\bf U}\right] +{\bf D}_0\right)\nonumber\\
\leq &  \exp \big( g(\theta,\tau) \cdot \phi_\Omega \big),
\end{align}
where $\phi_\Omega:=\sum_{i=1}^I\left[\mu\left( {\bf U}_i\right)+1\right]^\tau-1$ with ${\bf U}_i = \mathop{\arg\max}\limits_{ k\in \Omega_i} \{ \mu ({\bf B}_k)\}$ and $g(\theta,K) \geq \max\{\theta,\theta^K\}$. The last inequality follows from (\ref{eq:diagonal.2}) and (\ref{eq:g_function}).

\end{IEEEproof}

{\subsection{Proof of Proposition \ref{prop:exp.bound2}} 

\begin{IEEEproof} It follows from Condition (C2) that, for any $\theta>0$,
\begin{equation*}
\mathbb{E} \left\{\mu\left(\sum_{k=1}^K{\bf X}_k\right)\right\} = \frac{1}{\theta}\mathbb{E} \left\{\mu\left(\sum_{k=1}^K\theta{\bf X}_k\right)\right\} 
\end{equation*}
According to Lemma \ref{lem:exp.bound}, we then have 
\begin{align*}
\mathbb{E} \left\{\mu\left(\sum_{k=1}^K{\bf X}_k\right)\right\} \leq&\min_{\theta>0} \left\{\frac{1}{\theta}\cdot \log \mathbb{E}{\rm e}^{\mu\left(\sum_{k=1}^K\theta {\bf X}_k\right)}\right\}\nonumber\\
\leq&\min_{0<\theta<\frac{3}{c}}\left\{ \frac{1 }{\theta} \cdot g_2(\theta;c) \cdot  \phi_{\widetilde{\Omega}} \right\} \nonumber\\%
\leq&\min_{ 0<\theta<\frac{3}{c}} \left\{ \frac{ \phi_{\widetilde{\Omega}}\cdot \alpha_2(c)}{\theta} +   \frac{3 \theta  \phi_{\widetilde{\Omega}}}{6-2c\theta}\right\}. %
\end{align*}
{By using a computer algebra system\footnote{https://www.wolframalpha.com}, the above minimization is achieved as 
\begin{equation*}
\phi_{\widetilde{\Omega}} \left(\sqrt{2  \alpha_2(c) } + \frac{c\alpha_2(c) }{3}\right),
\end{equation*}
when
\begin{equation*}
\theta = 
\left\{
\begin{array}{ll}
  \frac{ 9\sqrt{2 \alpha_2(c)} +  6c \alpha_2(c)}{2c^2 \alpha_2(c) - 9},& \mbox{if $c^2\alpha_2(c)>\frac{9}{2}$;}     \\
  \frac{9\sqrt{2  \alpha_2(c)}-6c \alpha_2(c)}{9-2c^2 \alpha_2(c)},&    \mbox{if $c^2 \alpha_2(c)<\frac{9}{2}$.}
\end{array}
\right.
\end{equation*}}
This completes the proof.
\end{IEEEproof}}

{
\subsection{Proof of Theorem \ref{thm:ran.approx}}

\begin{IEEEproof} Since $(\frac{a}{K}+1)^{n}-1 \leq \frac{(a+1)^{n}-1}{K}$, $\forall a>0,n\in\mathbb{N}$, it follows from \eqref{eq:exp.bound3} that 
\begin{align}\label{eq:ran.approx.pr3}
\mathbb{E} \mu( \widehat{{\bf R}}_K - {\bf B}) = \mathbb{E} \mu\left( \sum_{k=1}^K \frac{{\bf R}_k - {\bf B}}{K} \right) \leq \left(\sqrt{2  \alpha_2(c) } +  \frac{c \alpha_2(c)}{3}  \right) \frac{\phi_{\widetilde{\Omega}}  }{K},
\end{align}
where $\phi_{\widetilde{\Omega}} = \sum\limits_{i=1}^I\big( \big(\mu({\bf U}_i)+1\big)^{|\widetilde{\Omega}_i|}-1\big)$ with ${\bf U}_i = \mathop{\arg\max}\limits_{k\in\widetilde{\Omega}_i} \{\mu({\bf R}_k-{\bf B} ) \}$.
Let $u = \max\limits_{1\leq k\leq K} \{\mu({\bf R}_k-{\bf B} )\}$ and then we have 
\begin{equation}\label{eq:ran.approx.pr2}
\phi_{\widetilde{\Omega}} \leq I\cdot [(u+1)^2 -1].
\end{equation}
In the case that $u \leq  \sqrt{ 1+2\epsilon \mu( {\bf B} ) }-1$, the combination of \eqref{eq:ran.approx.pr3} and \eqref{eq:ran.approx.pr2} leads to the result \eqref{eq:ran.approx2}. This completes the proof.
\end{IEEEproof}
}


\bibliographystyle{IEEEtran}
\bibliography{ref-rm}

\end{document}